\documentclass[letterpaper]{article} 
\usepackage[table]{xcolor} 

\usepackage{aaai25}  
\usepackage{times}  
\usepackage{helvet}  
\usepackage{courier}  
\usepackage[hyphens]{url}  
\usepackage{graphicx} 
\usepackage{natbib}  
\usepackage{caption} 
\usepackage{newfloat}
\usepackage{listings}
\DeclareCaptionStyle{ruled}{labelfont=normalfont,labelsep=colon,strut=off} 
\newcommand{\partitle}[1]{\smallskip \noindent \textbf{#1}.}
\usepackage{multirow}
\usepackage{graphicx}
\usepackage{makecell}
\usepackage{booktabs}
\usepackage{array}
\usepackage{amsmath}
\usepackage{amssymb}

\definecolor{lightgreen}{rgb}{0.85, 1.0, 0.85} 
\definecolor{lightblue}{rgb}{0.85, 0.85, 1.0} 
\definecolor{lightred}{rgb}{1.0, 0.85, 0.85} 
\definecolor{transwhite}{rgb}{1,1,1} 
\usepackage{pifont}
\usepackage{amssymb}
\usepackage{tikz}
\usepackage{threeparttable}
\definecolor{customblue}{HTML}{46075A}

\title{GenHMR: Generative Human Mesh Recovery}
\author {
    Muhammad Usama Saleem\textsuperscript{\rm 1}, 
    Ekkasit Pinyoanuntapong\textsuperscript{\rm 1}, 
    Pu Wang \textsuperscript{\rm 1}, 
    Hongfei Xue\textsuperscript{\rm 1}, 
    Srijan Das\textsuperscript{\rm 1}, 
    Chen Chen\textsuperscript{\rm 2}
}
\affiliations {
    \textsuperscript{\rm 1}University of North Carolina at Charlotte, Charlotte, NC, USA\\
    \textsuperscript{\rm 2}University of Central Florida, Orlando, FL, USA\\
    \{msaleem2, epinyoan, pwang13, hxue2, sdas24\}@charlotte.edu, chen.chen@crcv.ucf.edu
}

\usepackage{bibentry}

\begin{document}

\maketitle

\begin{abstract}
Human mesh recovery (HMR) is crucial in many computer vision applications; from health to arts and entertainment. HMR from monocular images has predominantly been addressed by deterministic methods that output a single prediction for a given 2D image. However, HMR from a single image is an ill-posed problem due to depth ambiguity and occlusions. Probabilistic methods have attempted to address this by generating and fusing multiple plausible 3D reconstructions,  but their performance has often lagged behind deterministic approaches.  In this paper, we introduce \textbf{GenHMR}, a novel generative framework that reformulates monocular HMR as an image-conditioned generative task, explicitly modeling and mitigating uncertainties in the 2D-to-3D mapping process. GenHMR comprises two key components: (1) \textbf{a pose tokenizer} to convert 3D human poses into a sequence of discrete tokens in a latent space, and (2)  \textbf{an image-conditional masked transformer} to learn the probabilistic distributions of the pose tokens, conditioned on the input image prompt along with randomly masked token sequence. During \textit{inference}, the model samples from the learned conditional distribution to iteratively decode high-confidence pose tokens, thereby reducing 3D reconstruction uncertainties.  To further refine the reconstruction, a 2D pose-guided refinement technique is proposed to directly fine-tune the decoded pose tokens in the latent space, which forces the projected 3D body mesh to align with the 2D pose clues. Experiments on benchmark datasets demonstrate that GenHMR significantly outperforms state-of-the-art methods. Project website can be found at \url{https://m-usamasaleem.github.io/publication/GenHMR/GenHMR.html}

\end{abstract}
\vspace{-7pt}

\begin{figure}[ht]
    \centering
    \includegraphics[width=0.85\linewidth]{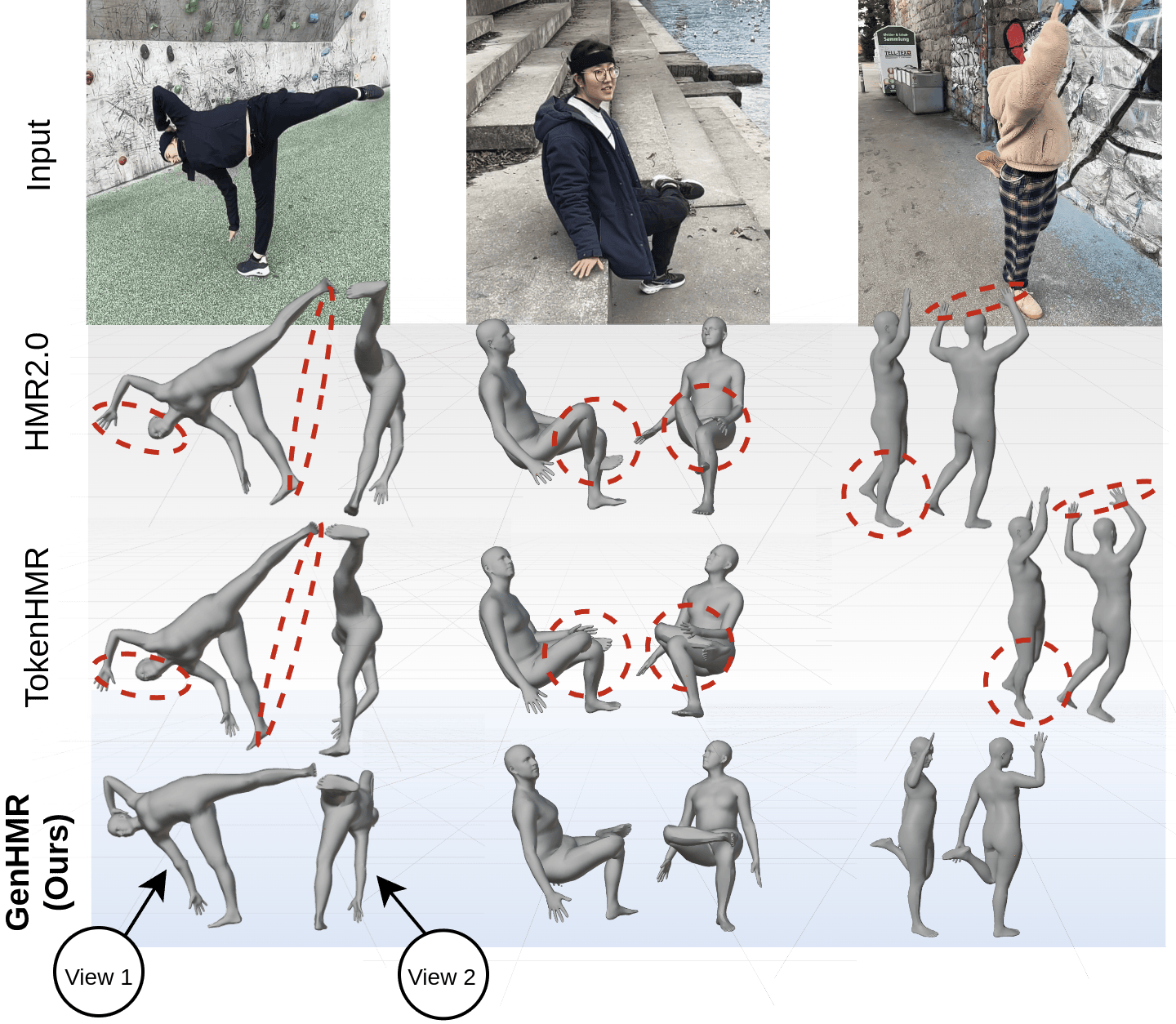}

    \caption{State of the art (SOTA) methods, HMR2.0 \cite{goel2023humans} and TokenHMR \cite{dwivedi2024tokenhmr}, leverage vision transformers to recover 3D human mesh from a single image. The errors, highlighted by red circles, reveal the limitations of these SOTA approaches in handling unusual poses or ambiguous scenarios. Our method, GenHMR, overcomes these challenges by explicitly modeling and mitigating uncertainties in the 2D-to-3D mapping process, resulting in more accurate and robust 3D pose reconstructions in such complex scenarios.}
    \label{fig:genhmr_sota}
\end{figure}

%

\section{Introduction}

Recovering 3D human mesh from monocular images is an essential task in computer vision, with applications spanning diverse fields, such as character animation for video games and movies,  metaverse, human-computer interaction, and sports performance optimization. However, recovering 3D human mesh from monocular images remains challenging due to inherent ambiguities in lifting 2D observations to 3D space, flexible body kinematic structures, complex intersections with the environment, and insufficient annotated 3D data ~\cite{tian2023recovering}. To address these challenges, recent efforts have been focusing on two methods: (1) deterministic HMR and (2) probabilistic HMR. Both methods face critical limitations.

Deterministic methods, as the dominant approach for HMR, are designed to produce a single prediction for a given 2D image. These methods estimate the shape and pose parameters of 3D  body model either by regressing from 2D image features extracted from deep neural networks ~\cite{choi2022learning, kocabas2021pare, kanazawa2018end} or by directly optimizing the parametric body model by fitting it to 2D image cues, such as 2D keypoints \cite{bogo2016keep,pavlakos2019expressive,xu2020ghum}, silhouettes \cite{omran2018neural}, and part segmentations \cite{lassner2017unite}. Recent deterministic HMR models, utilizing vision transformers as their backbone, have achieved state-of-the-art (SOTA) mesh reconstruction accuracy. However, despite such promising progress, these SOTA deterministic models face a critical limitation: they constrain neural networks to produce a single prediction hypothesis. This approach overlooks the inherent depth ambiguity in monocular images, which can result in multiple plausible 3D reconstructions that equally fit the same 2D evidence.

To mitigate this problem, several works have proposed probabilistic approaches that generate multiple predictions from a single image using various generative models, such as conditional variational autoencoders (CVAEs) \cite{sharma2019monocular, sohn2015learning} and diffusion models \cite{shan2023diffusion, holmquist2023diffpose}. However, this increase in diversity typically comes at the cost of accuracy because strategical aggregation of multiple solutions into a single accurate prediction is challenging due to the potential kinematic inconsistency among these 3D human mesh predictions. As a result, none of these multi-hypothesis probabilistic methods are competitive with the latest single-output deterministic models.

To overcome the limitations of existing methods, we introduce GenHMR, a novel generative framework for 3D human mesh recovery from a single 2D image. GenHMR is built on two key components: a pose tokenizer and an image-conditional masked transformer. The framework follows a two-stage training paradigm. In the first stage, the pose tokenizer is trained using Vector Quantized Variational Autoencoders (VQ-VAE) \cite{van2017neural}, which convert the continuous human pose (i.e., the rotations of skeletal joints) into a sequence of discrete tokens in a latent space, based on a learned codebook. In the second stage, a portion of the pose token sequence is randomly masked. The image-conditional masked transformer is then trained to predict the masked tokens by learning the conditional categorical distribution of each token, given the input image and the unmasked tokens.
This generative masking training allows GenHMR to learn a explicit probabilistic mapping from the 2D image to the human pose. Leveraging such  feature, we propose uncertainty-guided iterative sampling during inference, where the model decodes multiple pose tokens simultaneously in each iteration by sampling from the learned image-conditioned pose distributions. The tokens with low prediction uncertainties are kept and the others are re-masked and re-predicted in the next iteration. This feature allows GenHMR to  iteratively reduce 2D-to-3D mapping uncertainties and progressively correct the wrong joint rotations to improve the mesh reconstruction accuracy. To further refine the reconstruction quality, we propose a novel 2D pose-guided refinement technique, which directly optimizes the decoded pose tokens in the latent space, with an objective to force the projected 3D body mesh to align with the 2D pose clues. Our contributions are summarized as follows:

\begin{itemize}
    \item We introduce GenHMR, a novel generative framework for accurate HMR from a single image, which largely departs from existing deterministic and probabilistic methods in terms of model architecture, training paradigm and inference process.

    \item We leverage generative masking training to learn intricate image-conditioned pose distributions, thus effectively capturing the 2D-to-3D mapping uncertainty.
    \item We propose a novel iterative inference strategy, incorporating uncertainty-guided sampling followed by 2D pose-guided refinement to progressively mitigate HMR errors.
    \item We demonstrate through extensive experiments that GenHMR outperforms SOTA methods on standard datasets, including Human3.6M in the controlled environment, and 3DPW and EMDB for in-the-wild scenarios. For both cases,  GenHMR could lead to 20 - 30 \%   error reduction (in terms of MPJPE) compared with SOTA methods. Qualitative results shows that GenHMR is robust to ambiguous image observations (Fig. \ref{fig:genhmr_sota}). 
\end{itemize}

\section{Related Work}

\subsection{Deterministic HMR}
The field of Human Mesh Recovery (HMR) from monocular images has been primarily dominated by deterministic approaches, which aim to generate a single output for a given 2D image. The early work mainly adopts optimization-based approaches to fit a parametric model human model such as SMPL \cite{loper2015smpl} to 2D image 
cues \cite{pavlakos2019expressive, kolotouros2019learning, lassner2017unite}. Later on, learning-based methods become more prevalent, which leverage CNNs to directly regress SMPL parameters from images \cite{choi2022learning,kocabas2021pare,kanazawa2018end} and videos \cite{cho2023video,kanazawa2019learning}. Recently, vision transformers \cite{alexey2020image} have been adopted for HMR tasks. For example, HMR 2.0 \cite{goel2023humans} and TokenHMR \cite{dwivedi2024tokenhmr} achieve the state-of-the-art mesh reconstruction accuracy by leveraging transformer's ability of modeling the long-range correlations to learn the dependencies of different human body parts in HMR tasks.  However, these deterministic methods are limited by their single-output nature, which fails to capture the inherent depth ambiguity in complex scenarios, leading to reconstruction errors when multiple plausible 3D interpretations exist.

\subsection{Probabilistic HMR}

To address the limitations of deterministic methods, various probabilistic models have been exploited to address the inherent uncertainty in the reconstruction process and enable the generation of multiple plausible 3D mesh predictions from a single 2D image. These methods include mixture density networks (MDNs) \cite{bishop1994mixture, li2019generating}, conditional variational autoencoders (CVAEs) \cite{sohn2015learning, pavlakos2018learning}, and normalizing flows \cite{wehrbein2021probabilistic, kolotouros2021probabilistic}. Recent advancements in diffusion-based HMR models, such as DDHPose \cite{cai2024disentangled}, Diff-HMR \cite{cho2023generative}, and D3DP \cite{shan2023diffusion}, have shown significant promise in generating diverse and realistic human meshes. However, these approaches face a great challenge in synthesizing multiple predictions into a single, coherent 3D pose. Current methods often rely on simplistic aggregation techniques, such as averaging all hypotheses \cite{wehrbein2021probabilistic,li2019generating,li2020weakly} or performing pose or joint-level fusion \cite{shan2023diffusion,cai2024disentangled}, which frequently result in kinematically inconsistent and suboptimal reconstructions.

Unlike existing probabilistic methods that model and sample the distributions of entire 3D body meshes, GenHMR leverages masked generative transformers to model the pose and rotation distribution of each individual skeletal joint, while also capturing the inherent interdependence among joints. This approach enables our model to iteratively perform fine-grained joint-level sampling, progressively reconstructing the human mesh from ambiguous 2D image observations. Our generative framework draws inspiration from the great success of masked language and image models in text, image, and video generation \cite{ghazvininejad2019mask,devlin2019bert,zhang2021m6,qian2020glancing, chang2022maskgit,ding2022cogview2,chang2023muse}. However, while generative language and image models focus on increasing sample diversity, GenHMR aims to minimize 3D mesh generation diversity and uncertainty, conditioned on 2D image prompts.

\begin{figure*}[ht]
    \centering
    \includegraphics[width=0.65\linewidth]{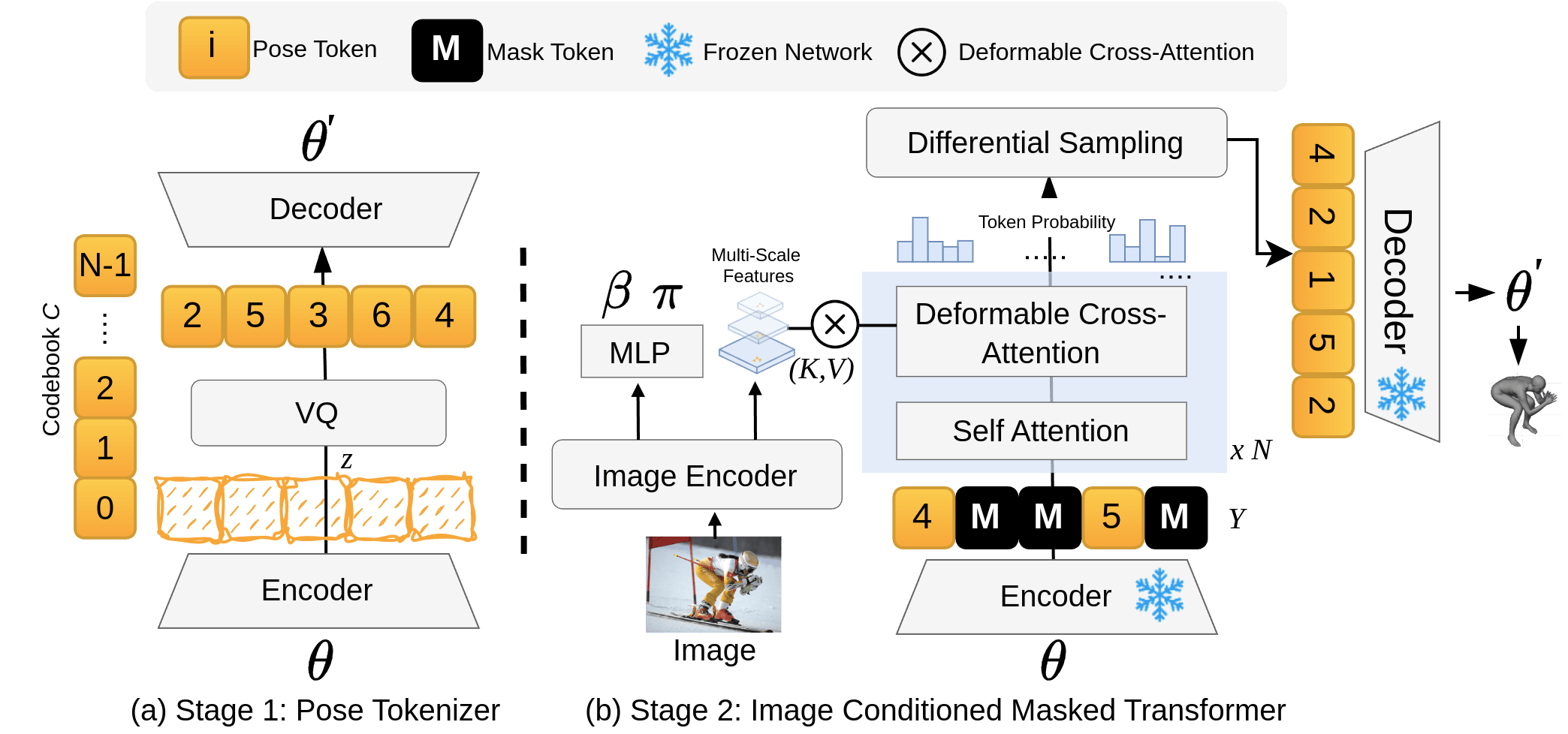}

\caption{GenHMR \textit{Training Phase}. GenHMR consists of two key components: (1) a \textbf{Pose Tokenizer} that encodes 3D human poses into a sequence of discrete tokens within a latent space, and (2) an \textbf{Image-Conditioned Masked Transformer} that models the probabilistic distributions of these tokens, conditioned on the input image and a partially masked token sequence.}
    \label{fig:genhmr_overview}
\end{figure*}

\begin{figure*}[ht]
    \centering
    \includegraphics[width=0.80\linewidth]{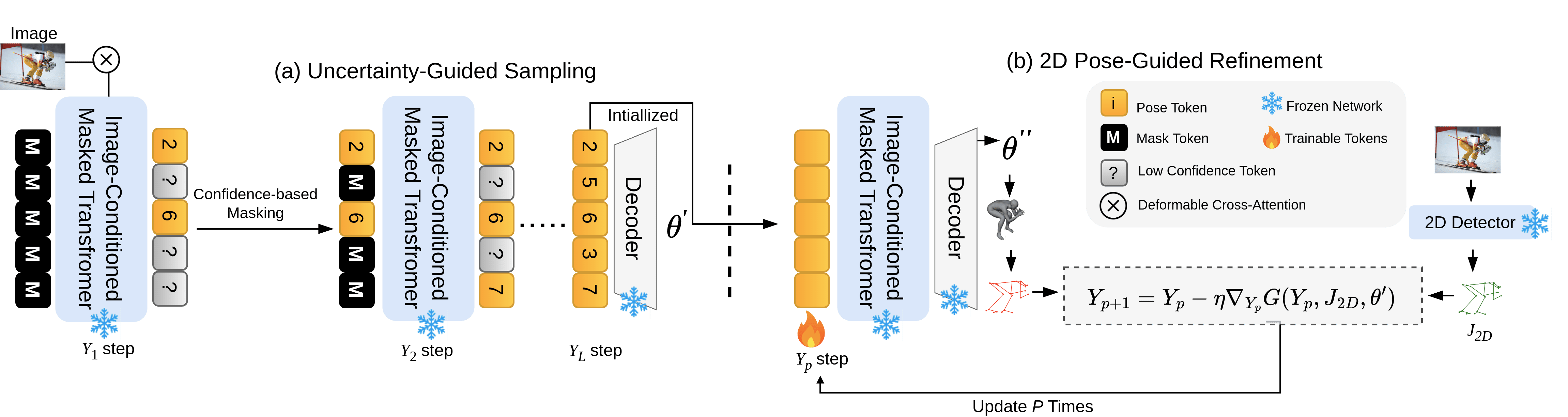}

\caption{Our \textit{inference strategy} comprises two key stages: (1) \textbf{Uncertainty-Guided Sampling}, which iteratively samples high-confidence pose tokens based on their probabilistic distributions, and (2) \textbf{2D Pose-Guided Refinement}, which fine-tunes the sampled pose tokens to further minimize 3D reconstruction uncertainty by ensuring consistency between the 3D body mesh and 2D pose estimates.}
    \label{fig:genhmr_inference}
\end{figure*}

\section{Proposed Method: GenHMR}
The goal of GenHMR is to achieve accurate 3D human mesh reconstructions from monocular images \(I\) by learning body pose \(\theta\), shape \(\beta\), and camera parameters \(T\). As shown in Figure \ref{fig:genhmr_overview}, GenHMR comprises two main modules: the pose tokenizer and the image-conditioned masked transformer. The pose tokenizer converts 3D human pose parameters \(\theta\) into a sequence of discrete pose tokens. The image-conditioned masked transformer predicts masked pose tokens based on multi-scale image features extracted by an image encoder. During inference, we use an iterative decoding process to predict high-confidence pose tokens, progressively refining predictions by masking low-confidence tokens and leveraging both image semantics and inter-token dependencies. To further enhance accuracy, a 2D pose-guided sampling strategy is proposed to optimize the pose queries by aligning the re-projected 3D pose with the estimated 2D pose. Additionally, the shape parameters \(\beta\) and weak perspective camera parameters \(T\) are directly regressed from the image features, completing the 3D human mesh reconstruction process.
\vspace{-5pt}

\subsection{Body Model} We utilize SMPL, a differentiable parametric body model \cite{loper2015smpl}, to represent the human body. The SMPL model encodes the body using pose parameters \(\theta \in \mathbb{R}^{24 \times 3}\) and shape parameters \(\beta \in \mathbb{R}^{10}\). The pose parameters \(\theta = [\theta_1, \ldots, \theta_{24}]\) include the global orientation \(\theta_1 \in \mathbb{R}^{3}\) of the whole body and the local rotations \([\theta_2, \ldots, \theta_{24}] \in \mathbb{R}^{23 \times 3}\) of the body joints, where each \(\theta_k\) represents the axis-angle rotation of joint \(k\) relative to its parent in the kinematic tree. Given these pose and shape parameters, the SMPL model generates a body mesh \(M(\theta, \beta) \in \mathbb{R}^{3 \times N}\), where \(N = 6890\) vertices. The body joints \(J \in \mathbb{R}^{3 \times k}\) are then defined as a linear combination of these vertices, calculated using \(J = MW\), where \(W \in \mathbb{R}^{N \times k}\) represents fixed weights that map vertices to joints.
\vspace{-5pt}

\subsection{Pose Tokenizer}

The goal of the pose tokenizer is to learn a discrete latent space for 3D pose parameters by quantizing the encoder's output into learned codebook $C$, as shown in Figure \ref{fig:genhmr_overview}(a). We leverage the VQ-VAE \cite{van2017neural} to pretrain the pose tokenizer. Specifically, given the SMPL pose parameters \(\theta\), we use a convolution encoder \(E\) to map the pose parameters \(\theta\) into a latent embedding \(z\). Each embedding \(z\) is then quantized to codes \(c \in C\) by finding the nearest codebook entry based on the Euclidean distance, described by $\hat{z}_i = \arg\min_{c_k \in C} \|z_i - c_k\|_2$. Then, the total loss function is defined as follows
\[
\mathcal{L}_{\text{vq}} = \lambda_{\text{re}}\mathcal{L}_{\text{re}} + \lambda_{\text{E}}\| \operatorname{sg}[z] - c \|_2 + \lambda_{\alpha}\|z - \operatorname{sg}[c]\|_2
\]
which consists of a SMPL reconstruction loss, a latent embedding loss and a commitment loss, where  $\lambda_{\text{re}}$, $\lambda_{\text{E}}$, and  $\lambda_{\alpha}$ are their respective weights.
 \(\operatorname{sg}[\cdot]\) represents the stop-gradient operator. To improve reconstruction quality, we employ an L1 loss $\mathcal{L}_{\text{re}} = \lambda_{\theta}\mathcal{L}_{\theta} + \lambda_{\text{V}}\mathcal{L}_{\text{V}} + \lambda_{\text{J}}\mathcal{L}_{\text{J}}$  to minimize the difference between the SMPL parameters and their ground-truth, including pose parameters $\theta$, mesh vertices $V$, and kinematic joints $J$. This tokenizer is optimized using a straight-through gradient estimator, with the codebooks being updated via exponential moving average and codebook reset, following the methodology outlined in \cite{esser2021taming,williams2020hierarchical}. 
 


\subsection{Image Conditioned Masked Transformer} 

The image conditioned masked transformer comprises two main components: the image encoder and the masked transformer decoder with multi-scale deformable cross attention.

\partitle{Image Encoder} Our encoder employs a vision transformer (ViT) to extract image features \cite{alexey2020image,dwivedi2024tokenhmr,goel2023humans}. We utilize the ViT-H/16 variant, which processes 16x16 pixel patches through transformer layers to generate feature tokens. Inspired by ViTDet \cite{alexey2020image}, we adopt a multi-scale feature approach by upsampling initial feature map from encoder to create a set of feature maps with varying resolutions. High-resolution feature maps capture fine-grained visual details (e.g., the presence and rotation of individual joints), while low-resolution feature maps preserve high-level semantics, e.g., the structure of the human skeleton. 

\partitle{Masked Transformer Decoder} Our decoder employs a  multi-layer transformer whose inputs are the pose token sequence obtained from the pose tokenizer. These pose tokens serve as the queries that  are cross-attended to the multi-scale feature maps from the image encoder. Since these feature maps are of high resolution, we adopt multi-scale deformable cross-attention to mitigate computational cost \cite{Zhu2020DeformableDD}. In particular, each pose token is only attended to a small set of sampling points around a reference point on multi-scale feature maps, regardless of the spatial size of the feature maps. The multi-scale deformable attention is expressed as:

\begin{equation*}
\begin{split}
\text{MSDA}(Y, \hat{p}_y, \{x^l\}_{l=1}^L) = \sum_{l=1}^L \sum_{k=1}^K A_{lyk} \cdot 
\mathbf{W} x^l \left( \hat{p}_y + \Delta p_{lyk} \right)
\end{split}
\end{equation*}
where $Y$ represents pose token queries, \(\hat{p}_y\) denotes learnable reference points, \(\Delta p_{lyk}\) are the learnable sampling offsets around these points, \(\{x^l\}_{l=1}^L\) are the multi-scale image features, \(A_{lyk}\) are the attention weights, and \(\mathbf{W}\) is a learnable weight matrix. The \texttt{[MASK]} token is a special-purpose token used in the masked transformer decoder. During training, it represents masked pose tokens, and the model learns to predict the actual tokens to replace them. During inference, \texttt{[MASK]} tokens serve as placeholders for pose token generation.

\vspace{-5pt}

\subsection{Training Strategy}

\partitle{Generative Masking} 
Given a pose token sequence $Y = [y_i]_{i=1}^L$ from the pose tokenizer where $L$ denotes the sequence length, our model is trained to reconstruct the pose token sequence, conditioned on the image prompt under random masking strategies. In particular, we randomly mask out \(m = \lceil \gamma(\tau) \cdot L \rceil\) tokens, where \(\gamma(\tau) \in [0, 1]\) is a masking ratio function with \(\tau\) following a uniform distribution \(U(0, 1)\). We adopt a cosine masking ratio function  \(\gamma(\tau) = \cos\left(\frac{\pi \tau}{2}\right) \) similar to the ones from generative  text-to-image models \cite{chang2022maskgit,chang2023muse}.   The masked tokens are replaced with learnable \texttt{[MASK]} tokens, forming the corrupted pose sequence \(Y_M\). The categorical distribution of each pose token, conditioned on corrupted sequence \(Y_M\) and image prompt $X$ is $p(y_i | Y_M, X)$, which explicitly models the uncertainty during the 2D-to-3D mapping process. The training objective is to minimize the negative log-likelihood of the pose token sequence prediction:

\begin{equation}
\mathcal{L}_{\text{mask}} = -\mathbb{E}_{Y \in \mathcal{D}} \left[ \sum_{\forall i \in [1, L]} \log p(y_i | Y_M, X) \right].
\end{equation}


\partitle{Training-time Differentiable Sampling}
The training loss, $\mathcal{L}_{\text{mask}}$, not only aids in capturing the uncertainty inherent in monocular human mesh recovery but also enforces accurate estimation of the pose parameter $\theta$ within the discrete latent space. However, prior research \cite{kanazawa2018end} indicates that, in addition to ensuring correct $\theta$ estimation, it is highly advantageous to incorporate an additional 3D loss between the predicted and ground-truth 3D joints, as well as a 2D loss between the projections of these predicted 3D joints and the ground-truth 2D joints. The challenge in incorporating these losses into generative model training lies in the need to convert pose tokens in the latent space into the pose parameter $\beta$ in the SMPL space. This conversion requires sampling the categorical distribution of pose tokens during training, which is non-differentiable. To overcome this challenge, we adopt the straight-through Gumbel-Softmax technique \cite{jang2016categorical}, which uses categorical sampling during the forward pass and employs differentiable sampling according to the continuous Gumbel-Softmax distribution during the backward pass, which can approximate the categorical distribution via temperature annealing. The final overall loss is $\mathcal{L}_{\text{total}} = \mathcal{L}_{\text{mask}} + \mathcal{L}_{\text{SMPL}} + \mathcal{L}_{\text{3D}} + \mathcal{L}_{\text{2D}}$, which combines pose token prediction loss (\(\mathcal{L}_{\text{mask}}\)), 3D loss ($\mathcal{L}_{\text{3D}}$), 2D loss ($\mathcal{L}_{\text{2D}}$), and SMPL parameter loss $\mathcal{L}_{\text{SMPL}}$ that minimize the shape and pose parameters in the SMPL space.

\vspace{-5pt}

\subsection{Inference Strategy} 

As shown in Fig. \ref{fig:genhmr_inference}, our inference strategy comprises two key stages:  (1) uncertainty-guided sampling, which iteratively samples high-confidence pose tokens based on their probabilistic distributions and (2) 2D pose-guided refinement, which fine-tunes the sampled pose tokens to further minimize 3D reconstruction uncertainty by ensuring the consistency between the 3D body mesh and 2D pose estimates. 

\partitle{Uncertainty-Guided Sampling} The sampling process begins with a fully masked sequence \(Y_1\) of length \(L\), where all tokens are initially set to \texttt{[MASK]}. The sequence is decoded over \(T\) iterations. At each iteration, \(t\), the masked tokens are decoded by performing stochastic sampling, where the tokens are sampled based on their prediction distributions $p(y_i | Y_M, X)$. After the token sampling, a certain number of tokens with low prediction confidences are re-masked and re-predicted in the next iteration. The number of tokens to be re-masked is determined by a masking schedule \(\lceil \gamma\left(\frac{t}{T}\right) \cdot L \rceil\), where \(\gamma\) is a decaying function of iteration $t$ that produces higher masking ratio in the early iterations when the prediction confidence is low, while yielding low masking ratio in the latter iterations when the prediction confidence increases as more context information becomes available from previous iterations. We adopt the cosine function for \(\gamma\) and the impact of other decaying functions is shown in the supplementary material.

\partitle{2D Pose-Guided Refinement} To further reduce uncertainties and ambiguities in the 3D reconstruction, we refine the pose tokens \(Y\), while keeping the whole network frozen, so that the 3D pose estimates are better aligned with 2D pose clues from off-the-shelf 2D pose detectors such as OpenPose \cite{cao2017realtime}. This optimization process is initialized by the pose tokens from uncertainty-guided sampling, and these tokens are then iteratively updated to minimize a composite guidance function \(G(Y_p, J_{\text{2D}}, \theta')\) that penalizes the misalignment of 3D and 2D poses along with regularization terms:
\begin{equation}
Y^+ = \arg\min_{Y_p} \left( \mathcal{L}_{2D}(J_{\text{3D}}') + \lambda_{\theta'} \mathcal{L}_{\theta'}(\theta') \right)
\label{eq:optimize_query}
\end{equation}
The term \( \mathcal{L}_{2D}(J_{\text{3D}}') \) ensures that the reprojected 3D joints \( J_{\text{3D}}' \) are aligned with the detected 2D keypoints \( J_{\text{2D}} \):
\begin{equation}
\mathcal{L}_{2D}(J_{\text{3D}}') = |\Pi(K(J_{\text{3D}}')) - J_{\text{2D}} |^2
\label{eq:2d_loss_conf}
\end{equation}
where \( \Pi(K(\cdot)) \) represents the perspective projection with camera intrinsics \( K \).  The regularization term \( \mathcal{L}_{\theta'}(\theta') \) ensures that the pose parameters \( \theta' \) remain close to the initial estimate, preventing excessive deviations and maintaining plausible human body poses. At each iteration \( p \), the pose embeddings \(Y_p\) are updated using the following gradient-based approach:
\begin{equation}
Y_{p+1} = Y_p - \eta \nabla_{Y_p} G(Y_p, J_{\text{2D}}, \theta')
\end{equation}
Here, \( \eta \) controls the magnitude of the updates to the pose embeddings, while \( \nabla_{Y_p} G(Y_p, J_{\text{2D}}, \theta') \) represents the gradient of the objective function with respect to the pose embeddings \(Y\) at iteration \( p \). This refinement process continues over \( P \) iterations and our experiments show that only a small number of iterations (5 to 10) is sufficient to yield satisfaroy enhancement.


\vspace{-7pt}

\section{Experiments}

\begin{table*}[h!]
\centering
\caption{Reconstructions Evaluated in 3D: Reconstruction errors (in mm) on the Human3.6M, 3DPW, and EMDB datasets. Lower values ($\downarrow$) indicate better performance. \underline{Underlined} results show the second-best performance in each column. \textcolor{blue}{Blue} indicates improvements of our method compared to the second-best method. \& -- \& means results are not reported.}

\scalebox{0.73}{
\begin{tabular}{ll|cc|ccc|ccc}
\toprule
\multicolumn{2}{c|}{} & \multicolumn{2}{c|}{Human3.6M} & \multicolumn{3}{c|}{3DPW} & \multicolumn{3}{c}{EMDB} \\
Methods & Venue & \makecell{PA-MPJPE \\ (↓)} & \makecell{MPJPE \\ (↓)} & \makecell{PA-MPJPE \\ (↓)} & \makecell{MPJPE \\ (↓)} & \makecell{MVE \\ (↓)} & \makecell{PA-MPJPE \\ (↓)} & \makecell{MPJPE \\ (↓)} & \makecell{MVE \\ (↓)} \\
\midrule
\multicolumn{10}{c}{\textbf{Deterministic HMR Methods}} \\
FastMETRO \cite{cho2022cross} & \textcolor{gray}{\textit{ECCV 2022}} & 33.7 & 52.2 & 65.7 & 109.0 & 121.6 & 72.7 & 108.1 & 119.2 \\
PARE \cite{kocabas2021pare} & \textcolor{gray}{\textit{ICCV 2021}} & 50.6 & 76.8 & 50.9 & 82.0 & 97.9 & 72.2 & 113.9 & 133.2 \\
Virtual Marker \cite{ma20233d} & \textcolor{gray}{\textit{CVPR 2023}} & - & - & 48.9 & 80.5 & 93.8 & - & - & - \\
CLIFF \cite{li2022cliff} & \textcolor{gray}{\textit{ECCV 2022}} & \underline{32.7} & 47.1 & 46.4 & 73.9 & 87.6 & 68.8 & 103.1 & 122.9 \\
HMR2.0 \cite{goel2023humans} & \textcolor{gray}{\textit{ICCV 2023}} & 33.6 & \underline{44.8} & \underline{44.5} & \underline{70.0} & \underline{84.1} & \underline{61.5} & \underline{97.8} & 120.1 \\
VQ HPS \cite{fiche2023vq} & \textcolor{gray}{\textit{ECCV 2024}} & - & - & 45.2 & 71.1 & 84.8 & 65.2 & 99.9 & \underline{112.9} \\
TokenHMR \cite{dwivedi2024tokenhmr} & \textcolor{gray}{\textit{CVPR 2024}} & 36.3 & 48.4 & 47.5 & 75.8 & 86.5 & 66.1 & 98.1 & 116.2 \\
\midrule
\multicolumn{10}{c}{\textbf{Probabilistic HMR Methods}} \\
Diff-HMR \cite{cho2023generative}$^\dagger$ & \textcolor{gray}{\textit{ICCV 2023}} & - & - & 55.9 & 94.5 & 109.8 & - & - & - \\
3D Multibodies \cite{biggs20203d}$^\dagger$ & \textcolor{gray}{\textit{NeurIPS 2020}} &  42.2 &  58.2 & 55.6 & 75.8 & - & - & - & - \\
ProHMR \cite{kolotouros2021probabilistic}$^\dagger$ & \textcolor{gray}{\textit{ICCV 2021}} & - & - & 52.4 & 84.0 & - & - & - & - \\
\rowcolor{gray!15} \textbf{GenHMR} & \textbf{\textit{Ours}} & \textbf{22.4} \textbf{\textcolor{blue}{\scriptsize{(10.3$\downarrow$)}}} & \textbf{33.5} \textbf{\textcolor{blue}{\scriptsize{(11.3$\downarrow$)}}} & \textbf{32.6} \textbf{\textcolor{blue}{\scriptsize{(13.8$\downarrow$)}}} & \textbf{54.7} \textbf{\textcolor{blue}{\scriptsize{(15.3$\downarrow$)}}} & \textbf{67.5} \textbf{\textcolor{blue}{\scriptsize{(17.3$\downarrow$)}}} & \textbf{38.2} \textbf{\textcolor{blue}{\scriptsize{(23.3$\downarrow$)}}} & \textbf{68.5} \textbf{\textcolor{blue}{\scriptsize{(31.4$\downarrow$)}}} & \textbf{76.4} \textbf{\textcolor{blue}{\scriptsize{(43.7$\downarrow$)}}} \\
\bottomrule
\end{tabular}
}
\label{tab:sota_results}
\begin{flushleft}
\footnotesize{$^\dagger$Results for existing probabilistic HMR methods are reported for 25 multiple hypotheses.}
\end{flushleft}
\end{table*}

\partitle{Datasets} We trained the pose tokenizer using the AMASS \cite{mahmood2019amass} standard training split and MOYO \cite{tripathi20233d}. For GenHMR, following prior work \cite{goel2023humans} and to ensure fair comparisons, we used standard datasets (SD): Human3.6M (H36M) \cite{ionescu2013human3}, COCO \cite{lin2014microsoft}, MPI-INF-3DHP \cite{mehta2017monocular}, and MPII \cite{andriluka20142d}.

\partitle{Evaluation Metrics} We evaluate GenHMR using the Mean Per Joint Position Error (MPJPE) and Mean Vertex Error (MVE) for 3D pose estimation accuracy. We also report the Procrustes-Aligned MPJPE (PA-MPJPE) to assess the alignment between predicted and ground-truth poses after rigid transformation. To quantify computational efficiency, we use Average Inference Time per Image (AITI) (s), similar to the average inference/optimization time per image reported by OpenPose \cite{cao2017realtime}.  However, AITI measures per-iteration processing time, enabling fine-grained analysis of our uncertainty-guided sampling and 2D Pose-guided refinement.  Lower values across all these metrics indicate better performance. GenHMR is tested on the Human3.6M testing split, following previous works \cite{kolotouros2019learning}. To evaluate GenHMR's generalization on challenging in-the-wild datasets with varying camera motions and diverse 3D poses, we test 3DPW \cite{von2018recovering} and EMDB \cite{kaufmann2023emdb} without training on them, ensuring a fair assessment on unseen data.

\vspace{-5pt}

\subsection{Comparison to State-of-the-art Approaches}

We evaluate our approach against a range of state-of-the-art deterministic and probabilistic HMR methods on the Human3.6M, 3DPW, and EMDB datasets, as detailed in Table \ref{tab:sota_results}. GenHMR consistently outperforms existing methods across key evaluation metrics—PA-MPJPE, MPJPE, and MVE—demonstrating its superior ability to produce accurate 3D reconstructions. A significant contributor to this success is GenHMR's capability to model and refine uncertainty throughout the reconstruction process, making it particularly effective in challenging scenarios involving complex poses and occlusions.

This capability enables GenHMR to not only achieve substantial performance gains on the controlled Human3.6M dataset but also to deliver impressive error reduction on the more challenging in-the-wild 3DPW and EMDB datasets. For example, GenHMR achieves a 25.2\% reduction in MPJPE on the Human3.6M dataset, a 21.8\% reduction on 3DPW, and a remarkable 29.9\% reduction on EMDB, compared to existing state-of-the-art methods. These consistent improvements across multiple datasets, even without training on in-the-wild datasets like 3DPW and EMDB, highlight GenHMR's effectiveness in addressing the complexities of real-world scenarios and delivering high-quality 3D reconstructions with unprecedented accuracy.

\begin{figure*}[ht]
    \centering
    \includegraphics[width=0.67\linewidth]{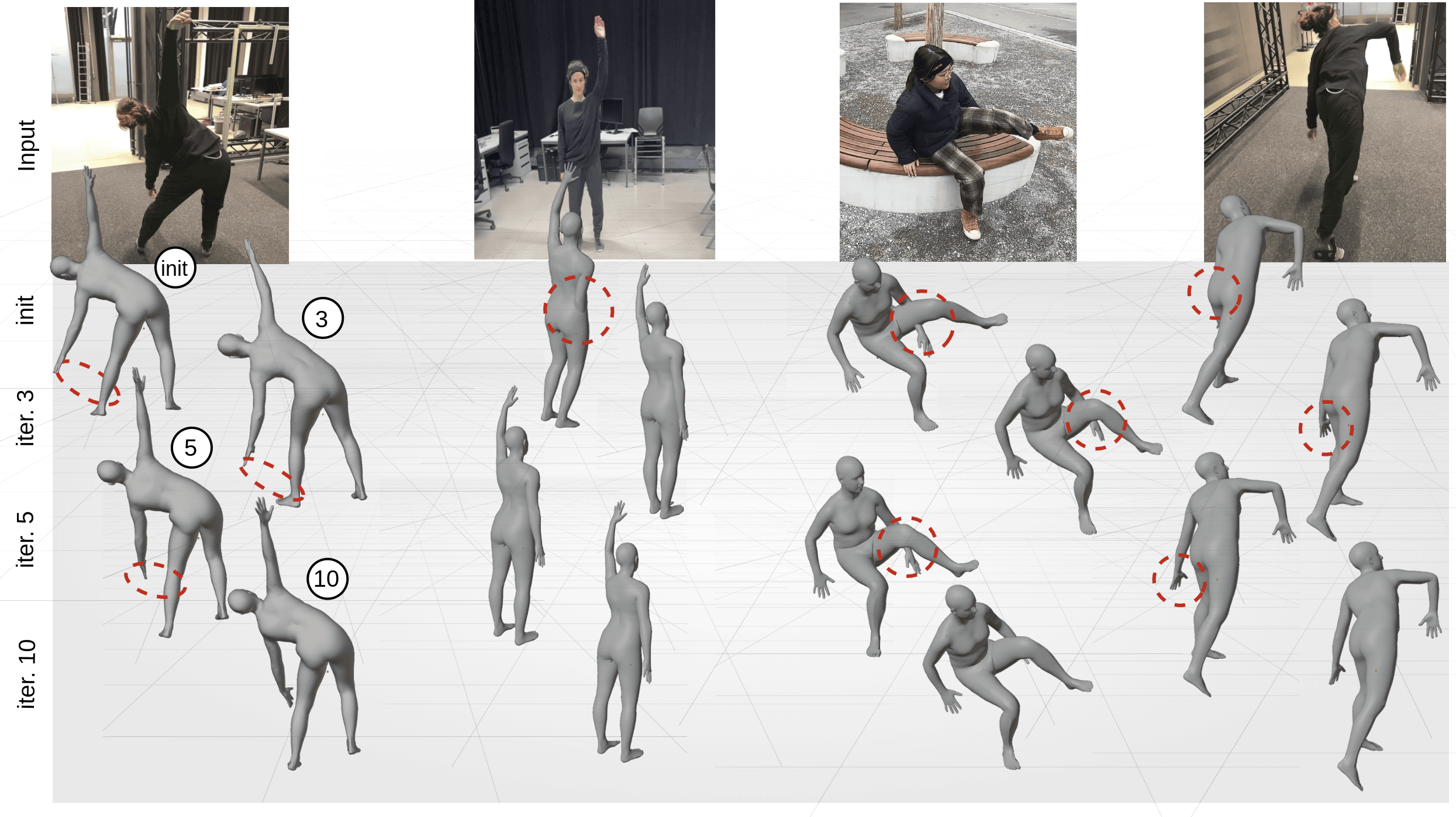}
    \caption{Impact of 2D Pose-Guided Refinement on 3D pose reconstruction. Red circles highlight areas of errors after each refinement iteration, which showcases how the method progressively refines these poses. By fine-tuning pose tokens to align the 3D pose with 2D detections, our method iteratively reduces uncertainties and improves accuracy. Significant improvements are seen in the early iterations, with errors largely minimized at the 10th iteration. Note that the initial mesh comes from UGS.}
    \label{fig:genhmr_inference_iter}
\end{figure*}
\vspace{-5pt}

\subsection{Ablation Study}

The key to GenHMR’s effectiveness lies in its mask modeling and iterative refinement techniques.  In this ablation study, we investigate how iterative refinement and mask-scheduling strategies influence the model's performance.  In the \textbf{Supplementary Material}, we provide extensive additional experimental results and visualizations that offer in-depth analyses of key factors contributing to GenHMR's performance. These include: (1) architectural components (pose tokenizer design, backbones, feature resolutions), (2) training strategies (masking scheduling, Gumbel-Softmax annealing, regularization via keypoint and SMPL losses in GenHMR), (3) inference techniques (speed-accuracy trade-offs in iterative refinement stages), (4) impact of training dataset size, and (5) model limitations.


\begin{table}[h!]
\caption{Impact of Iterations in Uncertainty-Guided Sampling}
\vspace{-7pt}
\centering
\scalebox{0.8}{
\begin{tabular}{ccccccc}
\toprule
& & \multicolumn{2}{c}{3DPW} & \multicolumn{2}{c}{EMDB} \\
\hline
\text{\# of iter.} & \text{AITI (sec)} & \makecell{MPJPE} & \makecell{MVE} & \makecell{MPJPE} & \makecell{MVE} \\
\toprule
1 & 0.032 & 73.5 & 84.2 & 92.2 & 104.5 \\
3 & 0.075 & 70.2 & 81.9 & 89.8 & 101.6 \\
5 & 0.102 & 68.1 & 77.5 & 88.2 & 99.5 \\
10 & 0.255 & 67.8 & 79.1 & 87.5 & 98.3 \\
15 & 0.321 & 67.6 & 78.5 & 87.1 & 96.6 \\
20 & 0.412 & 67.4 & 78.1 & 86.8 & 95.4 \\
\bottomrule
\end{tabular}
}
\label{tab:iterations_decoding}
\end{table}

\begin{table}[h!]
\centering
\caption{Impact of iterations on 2D Pose Guided Refinement.}
\vspace{-7pt}
\scalebox{0.8}{
\begin{threeparttable}
\begin{tabular}{ccccccc}
\toprule
& \multicolumn{1}{c}{H36M} & \multicolumn{2}{c}{3DPW} & \multicolumn{2}{c}{EMDB} & \multicolumn{1}{c}{AITI} \\
\cmidrule(lr){2-2} \cmidrule(lr){3-4} \cmidrule(lr){5-6} \cmidrule(lr){7-7}
\makecell{\# of iter.} & \makecell{MPJPE} & \makecell{MPJPE} & \makecell{MVE} & \makecell{MPJPE} & \makecell{MVE} & \makecell{(s)} \\
\midrule
UGS*  & 37.1 & 68.1 & 77.5 & 88.2 & 99.5 & 0.102 \\ 
1   & 36.2 & 66.5 & 77.0 & 86.5 & 96.5 & 0.154 \\ 
3   & 35.1 & 63.6 & 73.5 & 83.7 & 92.1 & 0.209 \\ 
5   & 34.3 & 60.2 & 70.4 & 79.1 & 87.5 & 0.253 \\ 
10  & 33.2 & 57.5 & 68.5 & 73.5 & 82.0 & 0.444 \\ 
20  & 33.5 & 54.7 & 67.5 & 68.5 & 76.4 & 0.638 \\
\bottomrule
\end{tabular}
\begin{tablenotes}
\footnotesize
\item[*] The first row UGS* reflects Uncertainty-Guided Sampling (UGS) with 5 iterations used to initialize the pose query \( Y \). AITI is obtained on a single mid-grade GPU (NVIDIA RTX A5000).
\end{tablenotes}
\end{threeparttable}
}
\label{tab:iterations_pose_sampling}
\end{table}

\partitle{Effectiveness of Uncertainty-Guided Sampling} The number of iterations during inference plays a critical role in balancing speed and accuracy in 3D pose estimation. As demonstrated in Table \ref{tab:iterations_decoding}, increasing the iterations generally enhances the accuracy of pose reconstructions, as reflected by improvements in MVE and MPJPE. For instance, on the 3DPW dataset, MVE decreases from 84.2 to 78.1, and MPJPE drops from 73.5 to 67.4 when the number of iterations increases from 1 to 20. However, beyond a certain threshold, such as after 10 iterations, the benefits of additional iterations diminish, with only marginal reductions in error. Notably, even with a smaller number of iterations, like 5, the model achieves significant accuracy improvements.

\partitle{Effectiveness of 2D Pose Guided Refinement} Fig. \ref{fig:genhmr_inference_iter} visualizes that 2D pose-guided refinement can iteratively improve 3D pose reconstruction accuracy particularly in scenarios with complex poses and occlusions. As shown in Table \ref{tab:iterations_pose_sampling}, significant accuracy gains occur early: on 3DPW, MVE drops by 11.04\% (77.0 mm to 68.5 mm) from 1 to 10 iterations, with minimal improvement to 67.5 mm at 20 iterations, where MVE improves by 12.34\% from 1 to 20 iterations. On EMDB dataset, MVE metric decreases by 15.02\% (96.5 mm to 82.0 mm) from 1 to 10 iterations, further reducing to 20.8\% (76.4 mm) at 20 iterations. These results underscore the method's effectiveness, though a trade-off with computational time is evident, as AITI increases from 0.154 seconds at 1 iteration to 0.638 seconds at 20 iterations. Notably, substantial gains are achievable with fewer iterations; at 5 iterations, MVE improves by 8.57\% on 3DPW (77.0 mm to 70.4 mm) and 9.3\% on EMDB (96.5 mm to 87.5 mm), while AITI remains manageable at 0.253 seconds. This highlights that 2D Pose-Guided refinement effectively balances accuracy and efficiency, making it practical for real-time human mesh recovery.

\begin{table}[h!]
\caption{Impact of masking ratio during training. These results were derived from the Uncertainty-Guided Sampling (UGS) stage with 5 iterations, where we evaluated the initial pose and shape estimate by iteratively refining the pose tokens.}
\vspace{-7pt}
\centering
\scalebox{0.8}{
\begin{tabular}{ccccc}
\toprule
& \multicolumn{2}{c}{3DPW} & \multicolumn{2}{c}{EMDB} \\
\cmidrule(lr){2-3} \cmidrule(lr){4-5}
Masking Ratio $\gamma(\tau)$ & \makecell{MPJPE} & \makecell{MVE} & \makecell{MPJPE} & \makecell{MVE} \\
\midrule
$\gamma(\tau \in \mathcal{U}(0, 1))$ & 68.1 & 77.5 & \textbf{88.2} & 99.5 \\$\gamma(\tau \in \mathcal{U}(0, 0.3))$ & 72.1 & 84.2 & 93.5 & 105.3 \\
$\gamma(\tau \in \mathcal{U}(0, 0.5))$ & 69.8 & 81.7 & 90.2 & 101.5 \\
$\gamma(\tau \in \mathcal{U}(0, 0.7))$ & \textbf{67.9} & \textbf{77.1} & 88.7 & \textbf{99.4} \\
\bottomrule
\end{tabular}
}
\label{tab:masking}
\end{table}

\partitle{Masking Ratio during Training} The ablation study on masking ratios during training underscores their influence on the accuracy of our generative model for human mesh recovery (HMR). We utilize a cosine-based masking ratio function \(\gamma(\tau)\), with \(\tau\) sampled from a uniform distribution, to randomly mask segments of the pose token sequence during training. This method induces varying levels of information loss, compelling the model to develop more robust reconstruction capabilities. The results reveal that broader masking ratios, such as \(\gamma(\tau \in \mathcal{U}(0, 0.7))\), lead to the most accurate reconstructions, demonstrated by the lowest MPJPE of 67.9 mm and a corresponding MVE of 77.1 mm on the 3DPW dataset. In contrast, narrower masking ratios, like \(\gamma(\tau \in \mathcal{U}(0, 0.3))\), result in higher error rates, with MPJPE rising to 72.1 mm and MVE to 84.2 mm, suggesting reduced generalization capabilities. These findings emphasize the pivotal role of selecting an appropriate masking ratio during training to achieve effective model generalization and precise 3D pose reconstruction.

\section{Conclusion}

In this paper, we introduced GenHMR, a novel generative approach to monocular human mesh recovery that effectively addresses the longstanding challenges of depth ambiguity and occlusion. By reformulating HMR as an image-conditioned generative task, GenHMR explicitly models and mitigates uncertainties in the complex 2D-to-3D mapping process. At its core, GenHMR consists of two key components: a pose tokenizer that encodes 3D human poses into discrete tokens within a latent space, and an image-conditional masked transformer that learns rich probabilistic distributions of these pose tokens. These learned distributions enable two powerful inference techniques: uncertainty-guided iterative sampling and 2D pose-guided refinement, which together produce robust and accurate 3D human mesh reconstructions. Extensive experiments on benchmark datasets demonstrate GenHMR's superiority over SOTA HMR methods.

\bibliography{aaai25}

\clearpage


\noindent \textbf{\huge Appendix}
\appendix
\subsection{Overview}

The appendix is organized into the following sections:

\begin{itemize}
    \item Implementation Details
    \item Data Augmentation
    \item Camera Model 
    \item Ablation for Pose Tokenizer 
    \item Training on Large-Scale Datasets
    \item Effectiveness of Training-time Differentiable Sampling
    \begin{itemize}
        \item Varying Temperatures in Cosine Annealing
        \item Impact of Annealing Durations
        \item Impact of Various Losses
    \end{itemize}

    \item Uncertainty Reduction Strategies at Inference 
    \begin{itemize}
        \item Confidence-based Masking
        \item Mask Scheduling Functions
        \item Token Sampling Strategies
    \end{itemize}
    
    \item Impact of Pose Tokenizer on GenHMR
    \item Ablation of Feature Resolutions
    \item Impact of Deformable Cross-Attention Layers
    \item Qualitative Results

\end{itemize}


\subsection{Implementation Details}
Our model, GenHMR, implemented using PyTorch, consists of two primary training stages: the pose tokenizer and the image-conditioned masked transformer. 

In the first stage, we train the pose tokenizer to learn discrete pose representations using mocap data from the AMASS \cite{mahmood2019amass} and MOYO \cite{tripathi20233d} datasets, which we converted from SMPL-H to SMPL format using SMPL-X instructions \cite{pavlakos2019expressive}. The resulting pose parameters $\theta \in \mathbb{R}^{24 \times 3}$ include both the global orientation $\theta_1 \in \mathbb{R}^{3}$ and the local rotations $[\theta_2, \ldots, \theta_{24}] \in \mathbb{R}^{23 \times 3}$ of the body joints. The pose tokenizer architecture comprises ResBlocks \cite{he2016deep} and 1D convolutions for both encoder and decoder, with a single quantization layer. We train the pose tokenizer for 200K iterations using the Adam optimizer, with a batch size of 512 and a learning rate of $1 \times 10^{-4}$. The loss weights for this stage are set to $\lambda_{\text{re}}=1.0$, $\lambda_{\text{E}}=0.02$, $\lambda_{\theta}=1.0$, $\lambda_{\text{V}}=0.5$, and $\lambda_{\text{J}}=0.3$. Based on the validation set reconstruction errors on the AMASS dataset, we select the final pose tokenizer model containing 96 tokens and a codebook size of $2048 \times 256$ for GenHMR training.

The second stage focuses on training the image-conditioned masked transformer, with the pose tokenizer from the first stage frozen to leverage the learned pose prior. To balance computational efficiency and mesh reconstruction accuracy, we utilize feature maps at 1x, 4×, and 8× resolutions. We adopt the Gumbel-Softmax operation for consistent end-to-end training, approximating the categorical distribution via temperature annealing. A cosine annealing schedule is used for the temperature parameter $\tau$, starting at $\tau_{\text{start}}=1.0$ and decreasing to $\tau_{\text{end}}=0.01$ with 50\% duration annealing. The final overall loss is $\mathcal{L}_{\text{total}} = \mathcal{L}_{\text{mask}} + \mathcal{L}_{\text{SMPL}} + \mathcal{L}_{\text{3D}} + \mathcal{L}_{\text{2D}}$, combining pose token prediction loss ($\mathcal{L}_{\text{mask}}$), 3D loss ($\mathcal{L}_{\text{3D}}$), 2D loss ($\mathcal{L}_{\text{2D}}$), and SMPL parameter loss $\mathcal{L}_{\text{SMPL}}$ that minimize the shape ($\beta$) and pose ($\theta$) parameters in the SMPL space. The loss weights for this stage are: $\lambda_{\text{mask}}=1.0$, $\lambda_{\text{SMPL}}=1.5 \times 10^{-3}$ (with $\lambda_{\theta}=1 \times 10^{-3}$ for pose and $\lambda_{\beta}=5 \times 10^{-4}$ for shape within $\mathcal{L}_{\text{SMPL}}$), $\lambda_{\text{3D}}=5 \times 10^{-2}$, and $\lambda_{\text{2D}}=1 \times 10^{-2}$. During inference in Uncertainty-Guided Sampling (UGS), we employ greedy sampling with $\text{top-k}=1$, selecting the most confident token predictions over a default of 5 iterations. We train the masked transformer on two NVIDIA RTX A6000 GPUs, using the Adam optimizer with a batch size of 48 and a learning rate of $1 \times 10^{-5}$.

\subsection{Data Augmentation}
In the first stage of training, the Pose Tokenizer benefits from incorporating prior information about valid human poses, which is crucial for the overall performance of GenHMR. To achieve this, we rotate the poses at varying degrees, enabling the model to learn a robust representation of pose parameters across different orientations. During the training of GenHMR, we further enhanced the model's robustness by applying random augmentations to both images and poses. These augmentations, including scaling, rotation, random horizontal flips, and color jittering, are designed to make the model more resilient to challenges such as occlusions and incomplete body information. Data augmentation is thus essential for improving the generalization and accuracy of the human mesh reconstruction process in GenHMR.

\subsection{Camera Model} Our method GenHMR employs a perspective camera model with a fixed focal length and an intrinsic matrix \(K \in \mathbb{R}^{3 \times 3}\). By simplifying the rotation matrix \(R\) to the identity matrix \(I_3\) and concentrating on the translation vector \(T \in \mathbb{R}^3\), we project the 3D joints \(J_{\text{3D}}\) to 2D coordinates \(J_{\text{2D}}\) using the formula \(J_{\text{2D}} = \Pi(K(J_{\text{3D}} + T))\), where \(\Pi\) represents the perspective projection with camera intrinsics \(K\). This simplification reduces parameter complexity and enhances the computational efficiency of our human mesh recovery pipeline.

\subsection{Ablation for Pose Tokenizer } In Tables \ref{tab:tokenizer_cb} and \ref{tab:tokenizer_tokens}, we present an extensive ablation study examining the design choices of the pose tokenizer on the AMASS test set and MOYO validation set, respectively. To assess out-of-distribution performance, most experiments are conducted using models trained exclusively on AMASS, while the final pose tokenizer employed in GenHMR was trained on both datasets. The findings underscore that the codebook size exerts a more substantial influence on performance than the number of pose tokens. Specifically, increasing the codebook size from 1024 to 4096 resulted in a reduction of Mean Vertex Error (MVE) by 3.3 mm (from 9.2 to 5.9) on AMASS and by 6.1 mm (from 14.5 to 8.4) on MOYO. Although increasing the number of tokens from 48 to 384 also enhanced performance (Table \ref{tab:tokenizer_tokens}), we selected a codebook size of 2048 × 256 with 96 tokens for our final model, optimizing the trade-off between performance and computational efficiency.

\begin{table}[h!]
\caption{Impact of Codebook Size (Pose Tokens = 96) on Pose Tokenizer.}
\vspace{-7pt}
\centering
\scalebox{0.9}{
\begin{tabular}{ccccc}
\toprule
 & \multicolumn{2}{c}{AMASS} & \multicolumn{2}{c}{MOYO} \\ \cmidrule(lr){2-3} \cmidrule(lr){4-5}
\makecell{\# of code $\,\times\,$ \\ code dimension} & \makecell{MPJPE} & \makecell{MVE} & \makecell{MPJPE} & \makecell{MVE} \\ 
\toprule
1024 $\,\times\,$ 256 & 10.5 & 9.2 & 16.2 & 14.5 \\
2048 $\,\times\,$ 128 & 8.9 & 8.1 & 13.9 & 12.7 \\
2048 $\,\times\,$ 256 & 8.5 & 7.8 & 13.4 & 12.2 \\
4096 $\,\times\,$ 256 &  \textbf{6.1} &  \textbf{5.9} &  \textbf{9.0} &  \textbf{8.4} \\
\bottomrule
\end{tabular}
}
\label{tab:tokenizer_cb}
\end{table}

\begin{table}[h!]
\caption{Impact of number of Pose Tokens (Codebook = 2048 $\times$ 256) on Pose Tokenizer}
\vspace{-7pt}
\centering
\scalebox{0.85}{
\begin{tabular}{ccccc}
\toprule
 & \multicolumn{2}{c}{AMASS} & \multicolumn{2}{c}{MOYO} \\ \hline
Tokens & \makecell{MPJPE} & \makecell{MVE} & \makecell{MPJPE} & \makecell{MVE} \\ 
\toprule
48 & 11.1 & 10.5 & 16.7 & 15.3 \\
96 & 8.5 & 7.8 & 13.4 & 12.2 \\
192 & 7.8 & 6.9 & 11.5 & 10.8 \\
384 & \textbf{7.1} & \textbf{6.0} & \textbf{10.7} & \textbf{10.2} \\
\bottomrule
\end{tabular}
}
\label{tab:tokenizer_tokens}
\end{table}

\subsection{Training on Large-Scale Datasets} 
To enhance GenHMR's real-world performance and generalization, we expanded our training data to include diverse datasets. We incorporated in-the-wild 2D datasets (ITW) such as InstaVariety \cite{kanazawa2019learning}, AVA \cite{gu2018ava}, and AI Challenger \cite{wu2017ai} with their pseudo ground truth (p-GT), and integrated the synthetic dataset BEDLAM (BL) as suggested by TokenHMR \cite{dwivedi2024tokenhmr}. This approach exposes our model to varied poses, camera motions, and environments. For a fair comparison, we only compare GenHMR with state-of-the-art models using the same backbone and datasets. To evaluate generalization, we test on unseen 3DPW \cite{von2018recovering} and EMDB \cite{kaufmann2023emdb} datasets, assessing performance on challenging in-the-wild scenarios.

As shown in Table \ref{tab:large_scale_dataset_results}, GenHMR consistently achieves the lowest error rates across all datasets and metrics, highlighting its superiority in 3D human mesh reconstruction. This success is particularly evident when GenHMR is trained on standard datasets (SD) combined with ITW, where it outperforms leading models like HMR2.0 \cite{goel2023humans} and TokenHMR \cite{dwivedi2024tokenhmr}, with significant reductions in MPJPE and MVE. The addition of BEDLAM further enhances its performance, as demonstrated on the challenging EMDB dataset, where GenHMR achieves an MPJPE of 67.5mm and an MVE of 74.8mm—reductions of 24.2mm and 34.6mm, respectively, compared to TokenHMR. These improvements stem from GenHMR's novel inference strategy, which combines Uncertainty-Guided Sampling and 2D Pose-Guided Refinement. This approach incrementally refines pose predictions by focusing on high-confidence tokens and aligning 3D estimates with 2D detections, ensuring accurate and consistent reconstructions. Consequently, GenHMR demonstrates enhanced generalization and robustness across diverse datasets, establishing itself as a leading solution for 3D human mesh reconstruction.

\begin{table*}[h!]
\centering
\caption{Training on Large-Scale Datasets Reconstruction errors (in mm) on the Human3.6M, 3DPW, and EMDB datasets. Lower values ($\downarrow$) indicate better performance. \underline{Underlined} results show the second-best performance in each column. \textcolor{blue}{Blue} indicates improvements of our method compared to the second-best method.}
\vspace{-7pt}
\scalebox{0.85}{
\begin{tabular}{ll|cc|ccc|ccc}
\toprule
\multicolumn{2}{c|}{} & \multicolumn{2}{c|}{Human3.6M} & \multicolumn{3}{c|}{3DPW} & \multicolumn{3}{c}{EMDB} \\
Training Datasets & Methods & \makecell{PA-MPJPE \\ (↓)} & \makecell{MPJPE \\ (↓)} & \makecell{PA-MPJPE \\ (↓)} & \makecell{MPJPE \\ (↓)} & \makecell{MVE \\ (↓)} & \makecell{PA-MPJPE \\ (↓)} & \makecell{MPJPE \\ (↓)} & \makecell{MVE \\ (↓)} \\
\toprule
\multirow{2}{*}{SD + ITW} & HMR2.0 & \underline{32.4} & \underline{50.0} & 54.3 & 81.3 & 94.4 & 79.3 & 118.5 & 140.6 \\
& TokenHMR & 33.8 & 50.2 & \underline{49.3} & \underline{76.2} & \underline{88.1} & \underline{67.5} & \underline{102.4} & \underline{124.4} \\
\rowcolor{gray!15} & \textbf{GenHMR} & \textbf{24.3} \textbf{\textcolor{blue}{\scriptsize{(8.1$\downarrow$)}}} & \textbf{37.8} \textbf{\textcolor{blue}{\scriptsize{(12.2$\downarrow$)}}} & \textbf{37.5} \textbf{\textcolor{blue}{\scriptsize{(11.8$\downarrow$)}}} & \textbf{58.6} \textbf{\textcolor{blue}{\scriptsize{(17.6$\downarrow$)}}} & \textbf{72.5} \textbf{\textcolor{blue}{\scriptsize{(15.6$\downarrow$)}}} & \textbf{44.5} \textbf{\textcolor{blue}{\scriptsize{(23.0$\downarrow$)}}} & \textbf{74.6} \textbf{\textcolor{blue}{\scriptsize{(27.8$\downarrow$)}}} & \textbf{82.8} \textbf{\textcolor{blue}{\scriptsize{(41.6$\downarrow$)}}} \\
\hline
\multirow{2}{*}{SD + ITW + BL} & HMR2.0 & \underline{28.5} & 47.7 & 47.4 & 77.4 & 88.4 & 62.8 & 99.3 & 120.7 \\
& TokenHMR & 29.4 & \underline{46.9} & \underline{44.3} & \underline{71.0} & \underline{84.6} & \underline{55.6} & \underline{91.7} & \underline{109.4} \\
\rowcolor{gray!15} & \textbf{GenHMR} & \textbf{22.1} \textbf{\textcolor{blue}{\scriptsize{(6.4$\downarrow$)}}} & \textbf{32.1} \textbf{\textcolor{blue}{\scriptsize{(14.8$\downarrow$)}}} & \textbf{31.0} \textbf{\textcolor{blue}{\scriptsize{(13.3$\downarrow$)}}} & \textbf{52.1} \textbf{\textcolor{blue}{\scriptsize{(18.9$\downarrow$)}}} & \textbf{65.6} \textbf{\textcolor{blue}{\scriptsize{(19.0$\downarrow$)}}} & \textbf{37.5} \textbf{\textcolor{blue}{\scriptsize{(18.1$\downarrow$)}}} & \textbf{67.5} \textbf{\textcolor{blue}{\scriptsize{(24.2$\downarrow$)}}} & \textbf{74.8} \textbf{\textcolor{blue}{\scriptsize{(34.6$\downarrow$)}}} \\
\bottomrule
\end{tabular}
}
\label{tab:large_scale_dataset_results}
\end{table*}

\section{Effectiveness of Training-time Differentiable Sampling}

\partitle{Varying Temperature Schedules in Cosine Annealing} In our study, we explored the impact of various temperature schedules on model performance using the Gumbel-Softmax estimator with cosine annealing. The temperature parameter (\(\tau\)) plays a crucial role in balancing exploration and exploitation during the learning process. We implemented a cosine annealing schedule to gradually reduce \(\tau\) from an initial value \(\tau_{\text{start}}\) to a final value \(\tau_{\text{end}}\) over \(T\) training iterations, following the equation:

\begin{equation}
\tau_t = \tau_{\text{end}} + \frac{1}{2} (\tau_{\text{start}} - \tau_{\text{end}}) \left(1 + \cos\left(\frac{t \pi}{T}\right)\right)
\end{equation}

This schedule allows for a smooth transition from broad exploration (higher \(\tau\)) to focused exploitation (lower \(\tau\)). The Gumbel-Softmax distribution, governed by \(\tau\), determines the probability of selecting codebook entries:

\begin{equation}
p_{m,k} = \frac{\exp\left(\frac{l_{m,k} + n_k}{\tau}\right)}{\sum_{j=1}^{K} \exp\left(\frac{l_{m,j} + n_j}{\tau}\right)}
\end{equation}
where \(l_{m,k}\) are logits and \(n_k\) are Gumbel noise samples. As \(\tau\) decreases, this distribution sharpens, leading to more deterministic predictions. 

Our experiments, detailed in Table \ref{tab:tau_ablation}, revealed that starting with a moderate temperature (\(\tau_{\text{start}} = 1.0\) or \(0.8\)) and annealing to a very low temperature (\(\tau_{\text{end}} = 0.01\)) yielded optimal performance, minimizing both MPJPE and MVE. This approach effectively balances initial exploration with subsequent exploitation. We found that higher starting temperatures (e.g., \(\tau_{\text{start}} = 2.0\)) led to slower convergence and increased errors due to excessive exploration, while lower starting temperatures (e.g., \(\tau_{\text{start}} = 0.5\)) accelerated convergence but reduced pose diversity, ultimately resulting in higher errors. Our findings underscore the importance of a carefully tuned temperature schedule in achieving both accuracy and generalization in accurate human mesh reconstruction.

\begin{table}[h!]
\caption{Impact of different temperature (\(\tau\)) schedules on the performance of human mesh recovery during training with the Gumbel-Softmax estimator using a cosine annealing schedule. Here, results are from \textbf{5 iterations of UGS} to evaluate initial  3D pose estimates at inference.}
\vspace{-7pt}
\centering
\scalebox{0.85}{
\begin{tabular}{cccccc}
\toprule
\multicolumn{2}{c}{Temperature (\(\tau\))} & \multicolumn{2}{c}{3DPW} & \multicolumn{2}{c}{EMDB} \\
\cmidrule(lr){1-2} \cmidrule(lr){3-4} \cmidrule(lr){5-6}
$\tau_{start}$  & $\tau_{end}$ & \makecell{MPJPE} & \makecell{MVE} & \makecell{MPJPE} & \makecell{MVE} \\
\midrule
2.0 & 0.01 & 90.1 & 99.5 & 110.2 & 120.5 \\
2.0 & 0.1  & 85.4 & 95.9 & 105.7 & 115.8 \\
1.5 & 0.01 & 78.3 & 88.2 & 96.5 & 107.3 \\
1.5 & 0.05 & 82.1 & 91.6 & 100.3 & 111.1 \\
1.2 & 0.01 & 74.5 & 83.2 & 92.0 & 103.5 \\
1.0 & 0.01 & 68.1 & \textbf{77.5} & \textbf{88.2} & 99.5 \\
0.8 & 0.01 & \textbf{68.0} & 78.2 & 88.8 & \textbf{99.1} \\
0.5 & 0.1  & 77.2 & 86.0 & 98.3 & 109.7 \\
\bottomrule
\end{tabular}
}
\label{tab:tau_ablation}
\end{table}

\partitle{Impact of Annealing Durations}  We conducted a comprehensive analysis of annealing duration effects on model performance, varied the annealing period as a percentage of total training iterations, employing a cosine schedule to gradually reduce the temperature ($\tau$). Our findings, presented in Table \ref{tab:annealing_ablation}, demonstrate the impact of annealing duration on the transition from smooth, differentiable sampling to more discrete, categorical sampling. This transition affects the model's ability to learn effective representations for human mesh recovery. Annealing durations spanning 50\% to 100\% of the training period consistently yield superior results, optimizing both accuracy and generalization in human mesh recovery. This range allows for an initial phase of broad exploration, gradually transitioning to focused exploitation of learned features. Shorter durations, such as 50\%, yield competitive results but slightly compromise generalization due to limited exploration time. Conversely, extended durations (e.g., 150\% or 200\%) maintain an exploratory state for too long, slightly diminishing the overall performance by delaying the transition to exploitation.

\begin{table}[h!]
\caption{Impact of different annealing durations on the performance of GenHMR using the Gumbel-Softmax estimator. Here, results reflect \textbf{5 iterations of UGS} to assess initial 3D pose estimates at inference }
\vspace{-7pt}
\centering
\scalebox{0.85}{
\begin{tabular}{ccccc}
\toprule
Annealing Duration & \multicolumn{2}{c}{3DPW} & \multicolumn{2}{c}{EMDB} \\
\cmidrule(lr){2-3} \cmidrule(lr){4-5}
 & MPJPE & MVE & MPJPE & MVE \\
\midrule
50\%  & \textbf{68.1} & \textbf{77.5} & \textbf{88.2} & 99.5 \\
75\%  & 80.5 & 89.4 & 99.7 & 110.5 \\
100\% & 68.9 & 78.2 & 88.9 & \textbf{98.8} \\
125\% & 70.0 & 79.0 & 89.5 & 99.2 \\
150\% & 72.5 & 81.0 & 90.8 & 100.7 \\
200\% & 75.2 & 84.1 & 94.3 & 105.0 \\
\bottomrule
\end{tabular}
}
\label{tab:annealing_ablation}
\end{table}

\partitle{Impact of Losses}  The results in Table \ref{tab:loss_impact} highlight the critical role of Training-Time Differentiable Sampling in enhancing GenHMR's performance, particularly when combined with key regularization losses. Incorporating all losses—\(L_{mask}\), \(L_{\theta}\), \(L_3D\), \(L_2D\), and \(\beta\)—yields optimal performance across all evaluation metrics on both the 3DPW and EMDB datasets. The 3D loss (\(L_3D\)) maintains structural integrity by aligning predicted 3D joints with ground truth, while the 2D loss (\(L_2D\)) addresses monocular reconstruction ambiguities by ensuring alignment between 3D projections and observed 2D keypoints. Notably, excluding either \(L_3D\) or \(L_2D\) leads to significant increases in errors, underscoring their vital role in producing accurate and plausible 3D reconstructions. These findings demonstrate that differentiable sampling not only enables seamless integration of these losses but also ensures their direct contribution to the overall accuracy and reliability of the model's predictions. Our analysis reveals that the combination of differentiable sampling and carefully chosen losses is crucial for achieving high-quality human mesh recovery. This approach allows the model to effectively learn from various constraints, resulting in more accurate and robust 3D reconstructions across different datasets and evaluation metrics.

\begin{table*}[h!]
\caption{Impact of different losses on MPJPE, PA-MPJPE, and MVE errors on the 3DPW and EMDB datasets. \textcolor{green}{\ding{51}} indicates inclusion, \textcolor{red}{\ding{55}} indicates exclusion, with deltas ($\Delta$) showing error differences when all losses are included. These results are from the \textbf{UGS stage with 5 iterations}, where we evaluated the initial 3D pose estimates }
\vspace{-7pt}
\centering
\scalebox{0.8}{
\begin{tabular}{ccccccccccccc}
\toprule
\multicolumn{5}{c}{Losses} & \multicolumn{3}{c}{3DPW} & \multicolumn{3}{c}{EMDB} \\
\cmidrule(lr){1-5} \cmidrule(lr){6-8} \cmidrule(lr){9-11}
$L_{mask}$ & $L_{\theta}$ & $L_3D$ & $L_2D$ & $\beta$ & PA-MPJPE (↓) & MPJPE (↓) & MVE (↓) & PA-MPJPE (↓) & MPJPE (↓) & MVE (↓) \\
\midrule
\textcolor{green}{\ding{51}} & \textcolor{green}{\ding{51}} & \textcolor{green}{\ding{51}} & \textcolor{green}{\ding{51}} & \textcolor{green}{\ding{51}} & 42.1 & 68.1 & 77.5 & 51.7 & 88.2 & 99.5 \\
\textcolor{green}{\ding{51}} & \textcolor{red}{\ding{55}} & \textcolor{red}{\ding{55}} & \textcolor{red}{\ding{55}} & \textcolor{green}{\ding{51}} & 45.5$_{\Delta 3.4}$ & 80.5$_{\Delta 12.4}$ & 92.5$_{\Delta 15.0}$ & 60.8$_{\Delta 9.1}$ & 98.8$_{\Delta 10.6}$ & 113.9$_{\Delta 14.4}$ \\
\textcolor{red}{\ding{55}} & \textcolor{green}{\ding{51}} & \textcolor{red}{\ding{55}} & \textcolor{red}{\ding{55}} & \textcolor{green}{\ding{51}} & 48.5$_{\Delta 6.4}$ & 82.0$_{\Delta 13.9}$ & 95.6$_{\Delta 18.1}$ & 61.7$_{\Delta 10.0}$ & 100.7$_{\Delta 12.5}$ & 118.0$_{\Delta 18.5}$ \\
\textcolor{red}{\ding{55}} & \textcolor{red}{\ding{55}} & \textcolor{green}{\ding{51}} & \textcolor{red}{\ding{55}} & \textcolor{green}{\ding{51}} & 57.9$_{\Delta 15.8}$ & 87.5$_{\Delta 19.4}$ & 146.4$_{\Delta 68.9}$ & 67.9$_{\Delta 16.2}$ & 110.9$_{\Delta 22.7}$ & 160.8$_{\Delta 61.3}$ \\
\textcolor{red}{\ding{55}} & \textcolor{red}{\ding{55}} & \textcolor{red}{\ding{55}} & \textcolor{green}{\ding{51}} & \textcolor{green}{\ding{51}} & 103.6$_{\Delta 61.5}$ & 1160.6$_{\Delta 1092.5}$ & 1167.7$_{\Delta 1090.2}$ & 110.7$_{\Delta 59.0}$ & 1180.8$_{\Delta 1092.6}$ & 1190.7$_{\Delta 1091.2}$ \\
\textcolor{green}{\ding{51}} & \textcolor{green}{\ding{51}} & \textcolor{red}{\ding{55}} & \textcolor{red}{\ding{55}} & \textcolor{green}{\ding{51}} & 45.1$_{\Delta 3.0}$ & 80.0$_{\Delta 11.9}$ & 91.5$_{\Delta 14.0}$ & 58.9$_{\Delta 7.2}$ & 97.6$_{\Delta 9.4}$ & 110.6$_{\Delta 11.1}$ \\
\textcolor{green}{\ding{51}} & \textcolor{green}{\ding{51}} & \textcolor{green}{\ding{51}} & \textcolor{red}{\ding{55}} & \textcolor{green}{\ding{51}} & \textbf{43.7}$_{\Delta \textbf{1.6}}$ & \textbf{78.5}$_{\Delta \textbf{10.4}}$ & \textbf{90.0}$_{\Delta \textbf{12.5}}$ & \textbf{56.9}$_{\Delta \textbf{5.2}}$ & \textbf{93.7}$_{\Delta \textbf{5.5}}$ & \textbf{106.9}$_{\Delta \textbf{7.4}}$ \\
\bottomrule
\end{tabular}
}
\label{tab:loss_impact}
\end{table*}

\subsection{Uncertainty Reduction Strategies at Inference }

\begin{figure*}[ht] 
    \centering
    \includegraphics[width=0.8\linewidth]{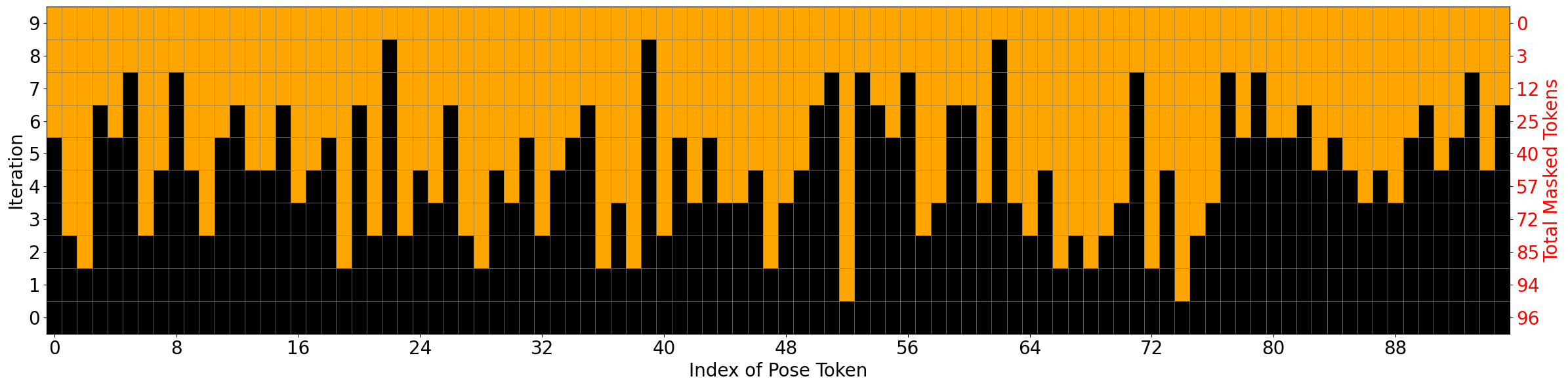}
\caption{ Visualization of mask tokens in each iteration. \tikz \draw[fill=black,draw=black] (0,0) rectangle (0.2,0.2); indicates [MASK] tokens, and \tikz \draw[fill=orange,draw=black] (0,0) rectangle (0.2,0.2); refers to unmasked tokens.}
\label{fig:genhmr_mask_infer_mask}
\end{figure*}

\begin{figure*}[ht] 
    \centering
    \includegraphics[width=0.8\linewidth]{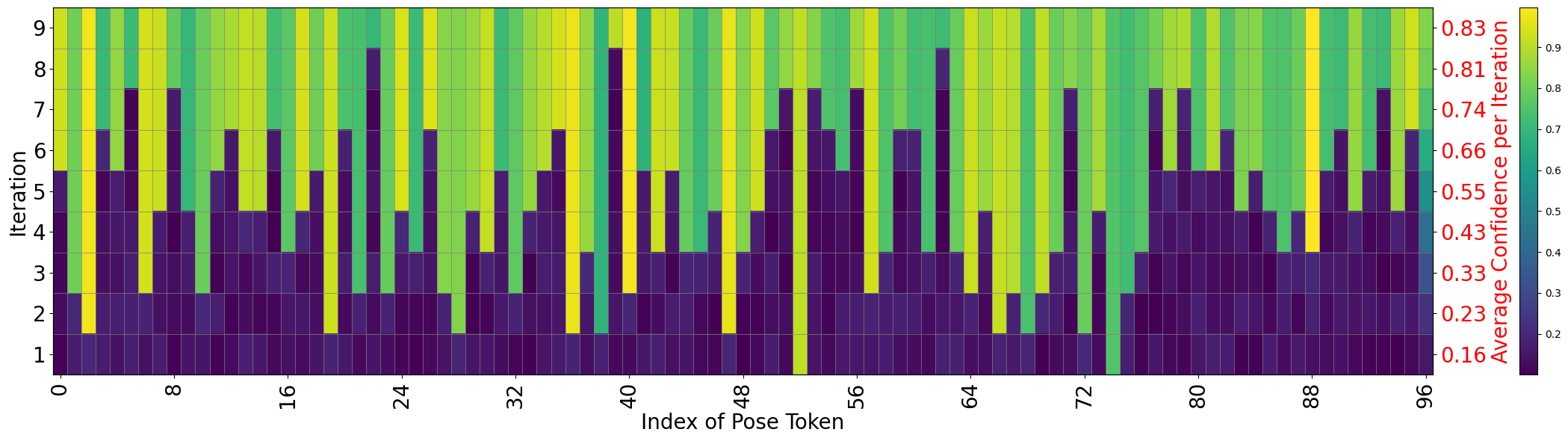}
\caption{Heatmap visualization of the Uncertainty-Guided Sampling Process. The heatmap illustrates the iterative decoding of a masked sequence over \(T\) iterations. The color gradient reflects prediction confidence, with \tikz \draw[fill=yellow,draw=black] (0,0) rectangle (0.2,0.2); representing high confidence and \tikz \draw[fill=customblue,draw=black] (0,0) rectangle (0.2,0.2); representing low confidence.}
\label{fig:genhmr_mask_infer_conf}
\end{figure*}

\partitle{Confidence-based Masking} During inference, we employ an Uncertainty-Guided Sampling process. This process is visualized in Figures \ref{fig:genhmr_mask_infer_mask} and \ref{fig:genhmr_mask_infer_conf}, both using the x-axis for 96 pose token indices (0-96) and the y-axis for iterations (0-9). The process begins with high uncertainty and progressively refines predictions. Figure \ref{fig:genhmr_mask_infer_mask} illustrates the masking pattern: black squares represent masked tokens, and orange squares unmasked tokens. Initially, most tokens are masked, reflecting the high uncertainty in early 3D pose estimates. As the model refines its predictions through iterations, the number of masked tokens decreases. Complementing this, Figure \ref{fig:genhmr_mask_infer_conf} shows prediction confidence levels: purple indicates low confidence, yellow high confidence. Early iterations display predominantly low confidence (purple), gradually shifting to higher confidence (yellow) in later iterations, mirroring the reduction in 2D-to-3D ambiguity. Notably, confidence typically increases for earlier tokens first, then spreads to later tokens, suggesting that initial context improves subsequent 3D estimates. This iterative approach enables GenHMR to transition from low-confidence to high-confidence predictions, leading to increasingly accurate and coherent human mesh recoveries while effectively reducing uncertainties in 2D-to-3D pose estimation.

\begin{figure}[ht] 
    \centering
    \includegraphics[width=0.9\linewidth]{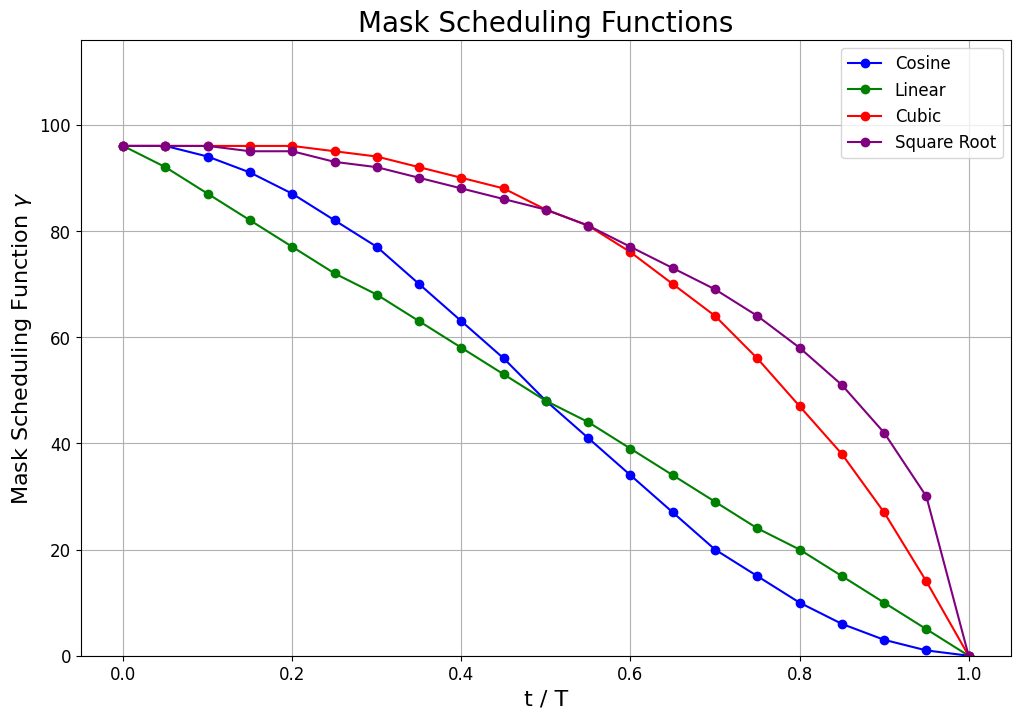}
\caption{Choices of Mask Scheduling Functions \(\gamma\left(\frac{t}{T}\right)\), and
number of iterations \(T\).}
\label{fig:genhmr_masking_funcs}

\caption{ \textit{}}

\end{figure}

\partitle{Mask Scheduling Function} The mask scheduling function determines how many tokens are re-masked at each iteration of the sampling process. We investigate four distinct masking schedule functions, illustrated in Figure \ref{fig:genhmr_masking_funcs}, to evaluate their influence on sequence generation. Let \(L\) denote the sequence length and \(T\) the total number of iterations. The number of tokens to be re-masked at iteration \(t\) is given by \(\lceil \gamma(t/T) \cdot L \rceil\), where \(\gamma\) is a decaying function of the normalized iteration time \(t/T\). The functions are as follows:

\begin{itemize}
    \item \textbf{Cosine Function}  yields a smooth S-curve, balancing gradual initial exploration, rapid mid-process transition, and refined final exploitation throughout decoding. 
    \[
    \gamma_{\text{cos}}\left(\frac{t}{T}\right) = \frac{1 + \cos\left(\frac{\pi t}{T}\right)}{2}
    \]

    \item \textbf{Linear Function} provides a constant decrease rate, maintaining a uniform balance between exploration and exploitation throughout the process.
    \[
    \gamma_{\text{linear}}\left(\frac{t}{T}\right) = 1 - \frac{t}{T}
    \]

    \item \textbf{Cubic Function}  function follows a concave curve, favoring extended initial exploration before rapidly transitioning to exploitation in later iterations. 
    \[
    \gamma_{\text{cubic}}\left(\frac{t}{T}\right) = 1 - \left(\frac{t}{T}\right)^3
    \]

    \item \textbf{Square Root Function} produces a convex curve, enabling rapid initial exploration followed by gradual convergence for extended fine-tuning in later iterations. 
    \[
    \gamma_{\text{sqrt}}\left(\frac{t}{T}\right) = \sqrt{1 - \left(\frac{t}{T}\right)^2}
    \]
\end{itemize}
Our analysis of masking functions in uncertainty-guided sampling (Table \ref{tab:genhmr_masking_funcs}) reveals significant performance variations across iterations and datasets, closely tied to each function's unique exploration-exploitation balance. The Cosine function consistently outperforms other methods on both 3DPW and EMDB datasets, achieving optimal performance at just 5 iterations. This suggests an efficient balance between initial exploration and rapid convergence to optimal solutions. While all functions improve with increased iterations, performance generally plateaus or slightly declines beyond 10-20 iterations. The Cubic and Linear functions, with their distinct decay patterns, reach peak performance at 20 iterations. In contrast, the Square Root function shows the least improvement, peaking early at 10 iterations, indicating a potentially premature transition from exploration to exploitation. These findings highlight how each function's unique masking strategy influences the model's ability to navigate the solution space effectively. The results underscore the critical importance of selecting appropriate masking functions for optimal performance in human mesh recovery tasks, emphasizing the need to balance thorough exploration of the pose space with efficient convergence to accurate solutions.

\begin{table}[h!]
\caption{Impact of Iterations and Masking Functions in Uncertainty-Guided Sampling}
\vspace{-7pt}
\centering
\scalebox{0.85}{
\begin{tabular}{cccccc}
\toprule
& \multicolumn{2}{c}{3DPW} & \multicolumn{2}{c}{EMDB} \\
\hline
\text{\# of iter.} & \makecell{MPJPE} & \makecell{MVE} & \makecell{MPJPE} & \makecell{MVE} \\
\toprule
\multicolumn{5}{c}{\textbf{Cosine}} \\
\hline
1  & 73.5 & 84.2 & 92.2 & 104.5 \\
3  & 70.2 & 81.9 & 89.8 & 101.6 \\
5  & \textbf{68.1} & \textbf{77.5} & \textbf{88.2} & \textbf{99.5} \\
10 & 67.8 & 79.1 & 87.5 & 98.3 \\
15 & 67.6 & 78.5 & 87.1 & 96.6 \\
20 & 67.4 & 78.1 & 86.8 & 95.4 \\
\midrule
\multicolumn{5}{c}{\textbf{Cubic}} \\
\hline
5  & 69.7 & \textbf{78.9} & 89.8 & 101.2 \\
10 & 69.1 & 79.1 & 89.0 & 100.1 \\
15 & 68.8 & 79.0 & 88.5 & 99.3 \\
20 & \textbf{68.6} & 79.2 & \textbf{88.2} & \textbf{98.7} \\
\midrule
\multicolumn{5}{c}{\textbf{Linear}} \\
\hline
5  & 71.6 & 81.2 & 91.7 & 103.1 \\
10 & 71.3 & \textbf{80.7} & 91.2 & 102.4 \\
15 & 71.0 & 80.9 & 90.8 & 101.3 \\
20 & \textbf{70.8} & 81.1 & \textbf{90.3} & \textbf{100.6} \\
\midrule
\multicolumn{5}{c}{\textbf{Square Root}} \\
\hline
5  & 73.4 & \textbf{82.8} & 93.5 & 104.9 \\
10 & \textbf{73.1} & 83.0 & \textbf{93.1} & \textbf{104.2} \\
15 & 73.3 & 83.3 & 93.3 & 104.6 \\
20 & 73.6 & 83.5 & 93.6 & 105.0 \\
\bottomrule
\end{tabular}
}
\label{tab:genhmr_masking_funcs}
\end{table}

\partitle{Token sampling strategies} The results demonstrate that the Top-k sampling strategy has a significant impact on GenHMR's performance as depicted in Table \ref{tab:top_k_sampling}. Specifically, using a Top-1 sampling approach yields the best results, with the lowest MPJPE and MVE across both the 3DPW and EMDB datasets (Table \ref{tab:top_k_sampling}). As the value of k increases, there is a noticeable decline in performance, with higher k values leading to increased MPJPE and MVE, indicating reduced accuracy in 3D pose estimation and mesh reconstruction. This suggests that restricting the model to the most confident token predictions (Top-1) is crucial for maintaining high precision in GenHMR, while broader sampling (higher k) introduces noise and uncertainty that degrade the overall performance of the estimated final pose.

\begin{table}[h!]
\caption{Impact of Top-k Sampling on GenHMR. Here, results are from \textbf{5 UGS iterations}, refining pose tokens to evaluate initial 3D pose estimates at inference.
}
\vspace{-7pt}
\centering
\scalebox{0.85}{
\begin{tabular}{ccccc}
\toprule
 & \multicolumn{2}{c}{3DPW} & \multicolumn{2}{c}{EMDB} \\ \hline
\makecell{Top-K} & \makecell{MPJPE } & \makecell{MVE} & \makecell{MPJPE )} & \makecell{MVE} \\ 
\toprule

1 & \textbf{68.1} & \textbf{77.5} & \textbf{88.2} & \textbf{99.5} \\

2 & 75.7 & 90.3 & 98.2 & 116.9 \\
5 & 85.1 & 98.6 & 108.6 & 123.1 \\
10 & 90.5 & 110.1 & 118.1 & 130.6 \\
20 & 98.9 & 120.6 & 130.9 & 140.3 \\
\bottomrule
\end{tabular}
}
\label{tab:top_k_sampling}
\end{table}

\subsection{Impact of Pose Tokenizer on GenHMR } The results from our experiments highlight the critical role of the Pose Tokenizer's design in the overall performance of GenHMR as shown in Table \ref{tab:stage2_cb} and  \ref{tab:stage2_tokens}. Specifically, our ablation studies on the codebook size demonstrate that an increase in codebook size initially enhances performance, as seen when moving from a $1024 \times 256$ to a $2048 \times 256$ configuration, with notable improvements in both MPJPE and MVE metrics across the 3DPW and EMDB datasets (Table \ref{tab:stage2_cb}). However, further expansion to a $4096 \times 256$ codebook leads to a decline in accuracy, indicating that while larger codebooks can provide richer pose representations, they may also introduce complexity that hinders 3D pose estimation in subsequent stages. Moreover, the number of tokens in the second stage critically impacts GenHMR's performance (Table \ref{tab:stage2_tokens}). A moderate token count (96) provides the best results, with lower (48) and higher (192, 384) counts leading to reduced accuracy. This highlights the need for an optimal balance between pose representation richness and the model's ability to effectively utilize this information for accurate 3D human mesh reconstruction.

\begin{table}[h!]
\caption{Impact of Codebook Size (Tokens = 96) on GenHMR. Here, results are from \textbf{5 iterations of UGS}, iteratively refining pose tokens to evaluate initial 3D pose estimates at inference.}
\vspace{-7pt}
\centering
\scalebox{0.85}{
\begin{tabular}{ccccc}
\toprule
 & \multicolumn{2}{c}{3DPW} & \multicolumn{2}{c}{EMDB} \\ \hline
\makecell{\# of code $\,\times\,$ \\ code dimension} & \makecell{MPJPE \\ (↓)} & \makecell{MVE \\ (↓)} & \makecell{MPJPE \\ (↓)} & \makecell{MVE \\ (↓)} \\ 
\toprule
1024 $\,\times\,$ 256 & 70.1 & 82.5 & 90.2 & 103.7 \\
2048 $\,\times\,$ 128 & 69.9 & 80.3 & 89.2 & 99.9 \\
2048 $\,\times\,$ 256 & \textbf{68.1} & \textbf{77.5} & \textbf{88.2} & \textbf{99.5} \\
4096 $\,\times\,$ 256 & 69.5 & 80.4 & 90.4 & 100.4 \\
\bottomrule
\end{tabular}
}
\label{tab:stage2_cb}
\end{table}

\begin{table}[h!]
\caption{Impact of Tokens (Codebook = 2048$\times$256) on GenHMR. Here, results are from \textbf{5 iterations of UGS}, iteratively refining pose tokens to evaluate initial 3D pose estimates at inference.}
\vspace{-7pt}
\centering
\scalebox{0.85}{
\begin{tabular}{ccccc}
\toprule
 & \multicolumn{2}{c}{3DPW} & \multicolumn{2}{c}{EMDB} \\ \hline
\makecell{\# of tokens} & \makecell{MPJPE \\ (↓)} & \makecell{MVE \\ (↓)} & \makecell{MPJPE \\ (↓)} & \makecell{MVE \\ (↓)} \\ 
\toprule
48 & 68.5 & 80.5 & 92.6 & 105.6 \\
96 & \textbf{68.1} & \textbf{77.5} & \textbf{88.2} & \textbf{99.5} \\
192 & 72.5 & 83.6 & 96.8 & 110.7 \\
384 & 82.5 & 91.5 & 101.4 & 135.4 \\
\bottomrule
\end{tabular}
}
\label{tab:stage2_tokens}
\end{table}

\subsection{Ablation of Feature Resolutions}
The ablation study on feature resolutions in GenHMR provides crucial insights into the efficacy of multi-scale feature representation for Human Mesh Recovery. Results, as shown in Table \ref{tab:feat_scale}, consistently demonstrate that increasing resolution from 1\(\times\) to 16\(\times\) significantly enhances accuracy, reducing MPJPE by 3.2 mm on 3DPW and 4.1 mm on EMDB. However, further increases yield diminishing returns, indicating an optimal balance between performance and computational efficiency at 16\(\times\) resolution. Crucially, the study reveals a symbiotic relationship between high and low-resolution features, with performance degrading when lower scales (1\(\times\) or 4\(\times\)) are omitted. This underscores the importance of GenHMR's multi-scale approach in capturing both fine-grained details and overall structure.

\begin{table}[h!]
\caption{Impact of feature resolutions on MPJPE error on 3DPW and EMDB datasets. \textcolor{green}{\ding{51}} indicates inclusion, \textcolor{red}{\ding{55}} indicates exclusion, with deltas ($\Delta$) showing error differences when we use all feature maps. Here, results are from \textbf{5 iterations of UGS}, iteratively refining pose tokens to evaluate initial pose estimate at inference.}
\vspace{-7pt}
\centering
\scalebox{0.85}{
\begin{tabular}{cccccc}
\toprule
\multicolumn{4}{c}{Feature Scale} & \multicolumn{2}{c}{MPJPE (↓)} \\
\cmidrule(lr){1-4} \cmidrule(lr){5-6}
1$\times$ & 4$\times$ & 8$\times$ & 16$\times$ & 3DPW & EMDB \\
\midrule
\textcolor{green}{\ding{51}} & \textcolor{green}{\ding{51}} & \textcolor{green}{\ding{51}} & \textcolor{green}{\ding{51}} & 68.3 & 88.3 \\
\textcolor{green}{\ding{51}} & \textcolor{red}{\ding{55}} & \textcolor{red}{\ding{55}} & \textcolor{red}{\ding{55}} & 71.3$_{\Delta 3.0}$ & 92.3$_{\Delta 4.0}$ \\
\textcolor{green}{\ding{51}} & \textcolor{green}{\ding{51}} & \textcolor{red}{\ding{55}} & \textcolor{red}{\ding{55}} & 69.8$_{\Delta 1.5}$ & 90.3$_{\Delta 2.0}$ \\
\textcolor{green}{\ding{51}} & \textcolor{green}{\ding{51}} & \textcolor{green}{\ding{51}} & \textcolor{red}{\ding{55}} & 68.8$_{\Delta 0.5}$ & 89.3$_{\Delta 1.0}$ \\
\textcolor{red}{\ding{55}} & \textcolor{green}{\ding{51}} & \textcolor{green}{\ding{51}} & \textcolor{green}{\ding{51}} & 68.8$_{\Delta 0.5}$ & 88.8$_{\Delta 0.5}$ \\
\textcolor{green}{\ding{51}} & \textcolor{red}{\ding{55}} & \textcolor{green}{\ding{51}} & \textcolor{green}{\ding{51}} & 69.3$_{\Delta 1.0}$ & 89.8$_{\Delta 1.5}$ \\
\textcolor{green}{\ding{51}} & \textcolor{green}{\ding{51}} & \textcolor{red}{\ding{55}} & \textcolor{green}{\ding{51}} & 69.8$_{\Delta 1.5}$ & 89.9$_{\Delta 1.6}$ \\
\bottomrule
\end{tabular}
}
\label{tab:feat_scale}
\end{table}

\subsection{Impact of Deformable Cross Attention Layers} 

The results show that the number of Deformable Cross Attention Layers is crucial for GenHMR's performance. Increasing the layers from 2 to 4 significantly improves MPJPE and MVE on the AMASS and MOYO datasets, enhancing 3D pose estimation and mesh reconstruction as shown in Table \ref{tab:cross_attention_layers}. However, adding more than 4 layers leads to diminishing returns and slight performance degradation. This indicates that 4 layers provide the optimal balance between complexity and performance, ensuring GenHMR achieves accurate and efficient 3D human mesh reconstruction.
\begin{table}[h!]
\caption{Impact of \# of Deformable Cross Attention Layers in GenHMR. Here, results are from \textbf{5 iterations of UGS}, iteratively refining pose tokens to evaluate initial  3D pose estimates at inference}
\vspace{-7pt}
\centering
\scalebox{0.85}{
\begin{tabular}{ccccc}
\toprule
 & \multicolumn{2}{c}{3DPW} & \multicolumn{2}{c}{EMDB} \\ \hline
\# of Deformable \\  Cross Attention \\  Layers & \makecell{MPJPE} & \makecell{MVE} & \makecell{MPJPE} & \makecell{MVE} \\ 
\toprule
2 & 70.5 & 85.9 & 94.8 & 107.0 \\
4 & 68.1 & 77.5 & 88.2 & \textbf{99.5} \\
6 & 69.5 & \textbf{77.1} & \textbf{88.1} & 100.2 \\
8 & 70.9 & 79.9 & 91.9 & 104.9 \\
\bottomrule
\end{tabular}
}
\label{tab:cross_attention_layers}
\end{table}

\subsection{Qualitative Results}

\begin{figure*}[ht] 
    \centering
    \includegraphics[width=0.9\linewidth]{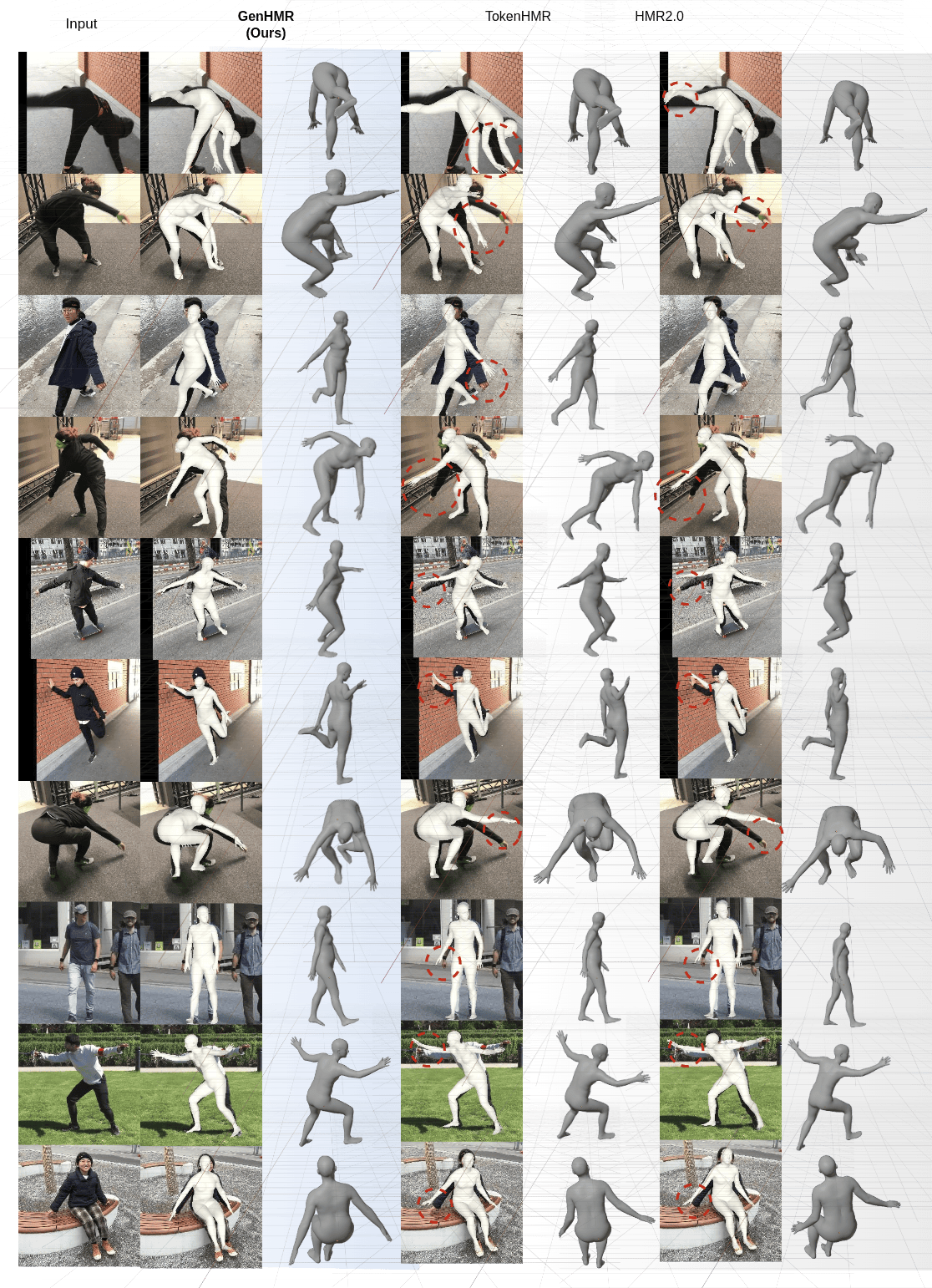}
\caption{State-of-the-art (SOTA) methods, such as HMR2.0 \cite{goel2023humans} and TokenHMR \cite{dwivedi2024tokenhmr}, utilize vision transformers to recover 3D human meshes from single images. However, the limitations of these SOTA approaches, particularly in dealing with unusual poses or ambiguous situations, are evident in the errors marked by red circles. Our approach, GenHMR, addresses these challenges by explicitly modeling and mitigating uncertainties in the 2D-to-3D mapping process, leading to more accurate and robust 3D pose reconstructions in complex scenarios. }
\label{fig:genhmr_sota_supp}
\end{figure*}

\begin{figure*}[ht] 
    \centering
    \includegraphics[width=1\linewidth]{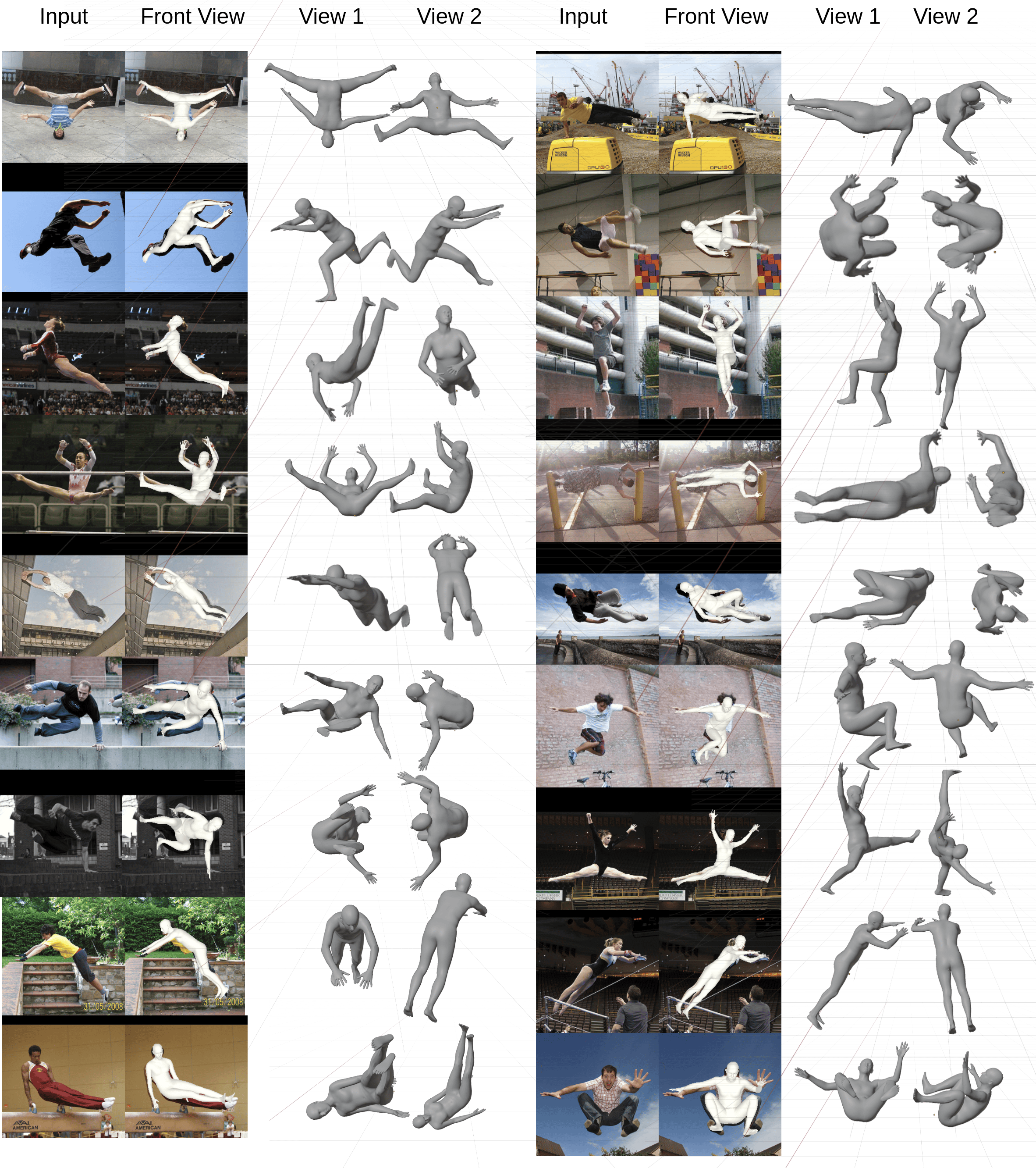}
\caption{Qualitative results of our approach on challenging poses from the LSP \cite{johnson2011learning} dataset. Results are directly from UGS.}
\label{fig:genhmr_chall_poses_3d}
\end{figure*}

\begin{figure*}[ht]
    \centering
    \includegraphics[width=0.67\linewidth]{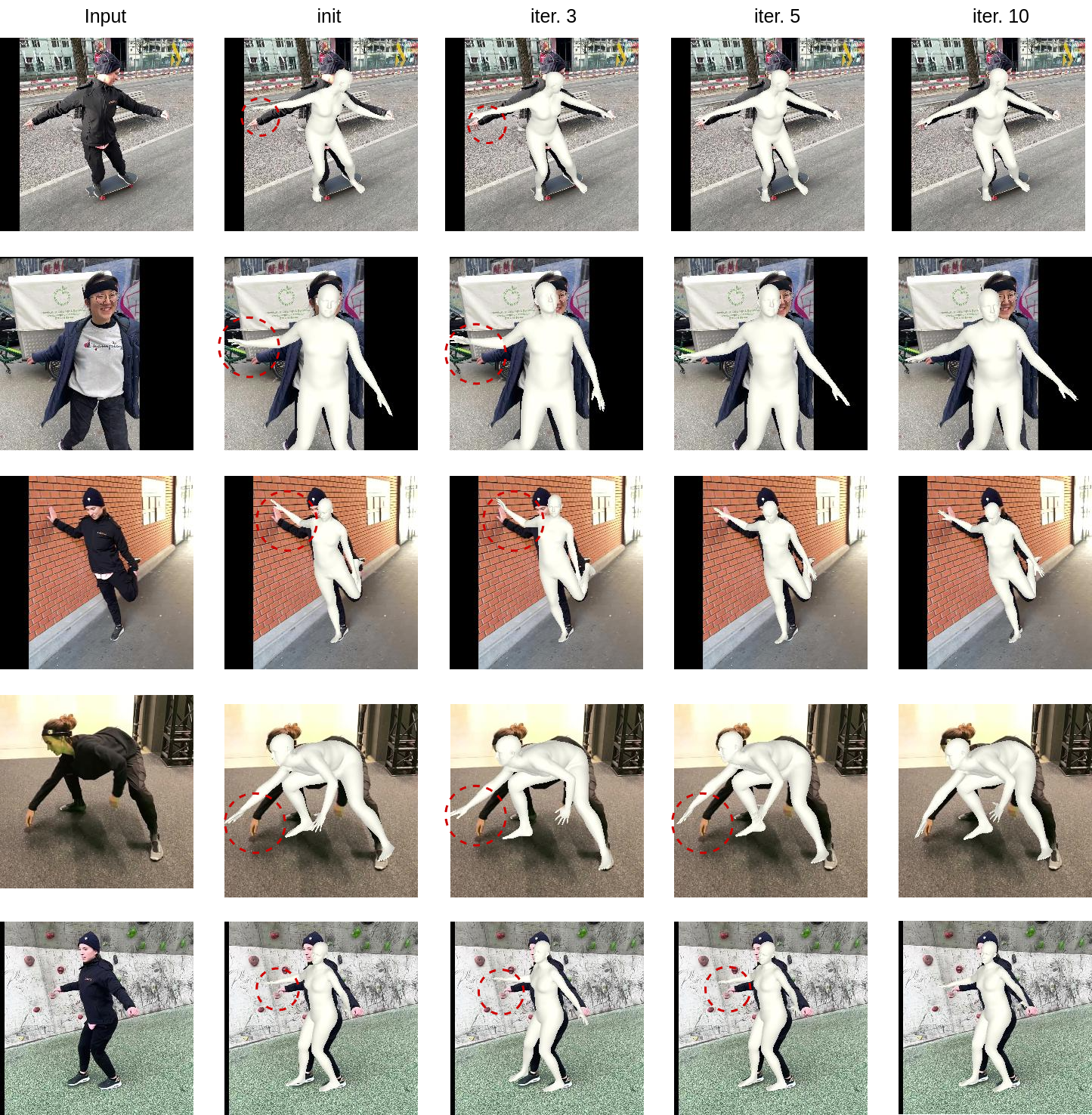}
    \caption{The effect of 2D Pose-Guided Refinement on 3D pose reconstruction. The red circles highlight error-prone areas after each refinement iteration, demonstrating how the method progressively corrects these errors. By fine-tuning the pose tokens to better align the 3D pose with 2D detections, our approach iteratively reduces uncertainties and enhances accuracy. Notable improvements are observed in the early iterations, with most errors significantly reduced by the 10th iteration. The initial mesh is derived from uncertainty-guided sampling.}
    \label{fig:genhmr_inference_iter_supp}
\end{figure*}

We present qualitative results of GenHMR in Figures \ref{fig:genhmr_sota_supp}, \ref{fig:genhmr_chall_poses_3d}, and \ref{fig:genhmr_inference_iter_supp}, demonstrating the model's robustness in handling extreme poses and partial occlusions. These results highlight the effectiveness of our approach, where reconstructions are well-aligned with the input images and remain valid when viewed from novel perspectives. A key factor contributing to this success is GenHMR’s explicit modeling and reduction of uncertainty during the 2D-to-3D mapping process. By iteratively refining pose estimates and focusing on high-confidence predictions, GenHMR is able to mitigate the challenges that typically hinder other state-of-the-art methods. This approach ensures more accurate and consistent 3D reconstructions, even in complex scenarios where traditional deterministic models often falter. Additionally, the refinement process, as illustrated in Figure \ref{fig:genhmr_inference_iter_supp}, plays a crucial role in aligning 3D outputs with 2D pose detections, further enhancing the model's ability to produce realistic and accurate meshes.

\begin{figure*}[ht] 
    \centering
    \includegraphics[width=1\linewidth]{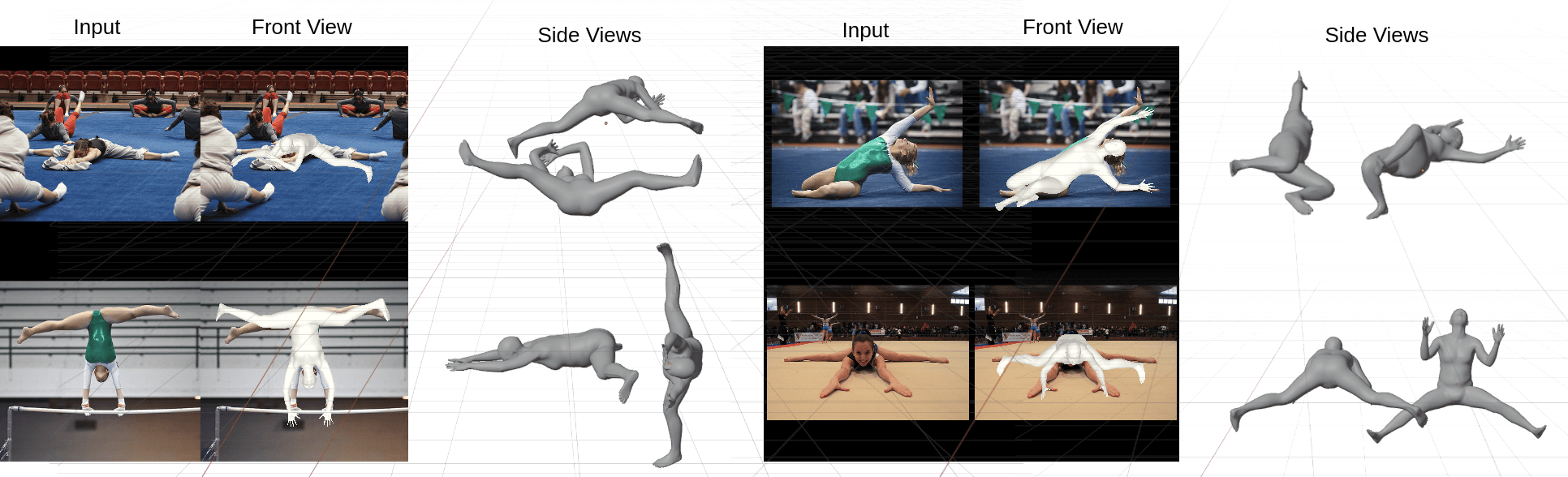}
\caption{Failure Cases of GenHMR in 3D Human Reconstruction: GenHMR often encounters errors when dealing with unusual body articulations and complex depth ordering of body parts. These challenges typically result in inaccurate 3D poses and non-valid outputs. The root of this limitation lies in the model's reliance on the SMPL parametric model, which may not fully capture the complexity of extreme or uncommon human poses.  }
\label{fig:genhmr_failure}
\end{figure*}





\end{document}